\documentclass[pdflatex,sn-mathphys-num]{sn-jnl}

\usepackage{graphicx}
\usepackage{multirow}
\usepackage{amsmath,amssymb,amsfonts}
\usepackage{amsthm}
\usepackage{mathrsfs}
\usepackage[title]{appendix}
\usepackage{xcolor}
\usepackage{textcomp}
\usepackage{manyfoot}
\usepackage{booktabs}

\usepackage{algorithm}
\usepackage{algpseudocode}
\usepackage{listings}
\usepackage{array}

\usepackage{subcaption}

\usepackage{stfloats}
\usepackage{url}
\usepackage{verbatim}

\usepackage{tcolorbox}
\tcbuselibrary{listingsutf8}

\usepackage{placeins}
\usepackage{cancel}
\usepackage{rotating}
\usepackage{arydshln}

\theoremstyle{thmstyleone}%
\theoremstyle{thmstyletwo}%

\theoremstyle{thmstylethree}%

\begin{document}
\title[Article Title]{LLM4Delay: Flight Delay Prediction via Cross-Modality Adaptation of Large Language Models and Aircraft Trajectory Representation}


\author[1]{\fnm{Thaweerath} \sur{Phisannupawong}}\email{petchthwr@kaist.ac.kr}

\author[1]{\fnm{Joshua J.} \sur{Damanik}}\email{joshuad@kaist.ac.kr}

\author*[1]{\fnm{Han-Lim} \sur{Choi}}\email{hanlimc@kaist.ac.kr}

\affil[1]{\orgdiv{Department of Aerospace Engineering}, \orgname{Korea Advanced Institute of Science and Technology}, \orgaddress{\city{Daejeon}, \postcode{34141}, \country{Republic of Korea}}}


\abstract{Flight delay prediction has become a key focus in air traffic management, as delays reflect inefficiencies in the system. This paper proposes LLM4Delay, a large language model-based framework for predicting flight delays from the perspective of air traffic controllers monitoring aircraft after they enter the terminal maneuvering area. LLM4Delay is designed to integrate textual aeronautical information, including flight data, weather reports, and aerodrome notices, together with multiple trajectories that model airspace conditions, forming a comprehensive delay-relevant context. By jointly leveraging comprehensive textual and trajectory contexts via instance-level projection, an effective cross-modality adaptation strategy that maps multiple instance-level trajectory representations into the language modality, the framework improves delay prediction accuracy. LLM4Delay demonstrates superior performance compared to existing ATM frameworks and prior time-series-to-language adaptation methods. This highlights the complementary roles of textual and trajectory data while leveraging knowledge from both the pretrained trajectory encoder and the pretrained LLM. The proposed framework enables continuous updates to predictions as new information becomes available, indicating potential operational relevance.}

\keywords{Air transportation, Artificial Intelligence, Data Analytics and Data Science, Flight Delay Prediction, Large Language Models}



\maketitle

\section{Introduction}\label{sec1}
The volume of air traffic grows in parallel with economic development, and can result in traffic density that exceeds the capacity of current air traffic management (ATM) systems, compromising both safety and operational efficiency~\cite{TBO, LpezLago2019}. The Federal Aviation Administration (FAA) defines delay as excess time incurred during flight operations and categorizes it into five phases: gate, taxi-out, en-route, terminal, and taxi-in~\cite{Mueller2002}. Delays can also be classified by cause, including carrier, weather, security, and airspace-related factors~\cite{Khanal2025}, with delays across phases and causes originating from different parts of the ATM system, each influenced by distinct factors. This paper focuses on delays after the terminal phase, of particular interest to air traffic controllers (ATCs) both as their direct operational responsibility and because updated arrival delay estimates facilitate coordination of post-landing ground services. ATCs monitor and integrate information across multiple modalities: textual information (flight schedules, weather reports, operational notices) provides operational context, while trajectory data captures aircraft movement, with multiple trajectories reflecting congestion. Integrating these contexts is essential for estimating delays based on their underlying causes. Although human-interpretable, such large-scale multimodal inference remains challenging for ATCs, motivating the development of predictive models.

Information in various modalities requires distinct handling techniques, and combining them necessitates effective data integration. Large language models (LLMs) have emerged as powerful tools for interpreting textual data and have demonstrated strong capability in understanding aviation-related text~\cite{notamevolve, Emmons2024}. However, trajectory data, as time series, are not in a format recent LLMs were trained to handle directly; therefore, integrating such data with LLMs requires additional adaptation techniques~\cite{timellm}. We propose LLM4Delay, a framework that integrates a pretrained LLM with a pretrained trajectory encoder by adapting learned instance-level semantics rather than sub-instance-level encodings. The framework treats all inputs as sequences of tokens to handle textual data and multiple trajectories with varying numbers of instances. This contrasts with existing frameworks that rely on fixed-size inputs, require rigid, handcrafted text parsers, and are limited to a single trajectory, thereby lacking detailed information about airspace and operational context that does not conform to predefined parsing rules. We formulate delay prediction as a multimodal regression problem leveraging both textual and trajectory data. This paper's contributions are as follows:

\begin{itemize}
    \item We introduce LLM4Delay, a multimodal framework that integrates aeronautical textual data and trajectory time-series data within a pretrained LLM to predict flight delays beyond the terminal phase.
    \item We propose instance-level projection, an effective cross-modality adaptation method that aligns multiple instance-level trajectory representations to language modality.
    \item We construct a publicly available multimodal delay-prediction dataset and demonstrate that integrating aeronautical textual and trajectory data provides complementary information, thereby improving prediction accuracy.
\end{itemize}

\section{Related Works}\label{sec2}
\subsection{Flight Delay Prediction}
Early flight delay prediction studies employed tabular-based approaches, extracting flight details (e.g., carrier, route, schedule, delay history) and weather data (e.g., visibility, wind, temperature) into structured tables as fixed-dimensional features for machine learning (ML) models, formulating the task as either classification or regression. For classification, predicting whether a delay will occur, studies have compared models such as Logistic Regression, Support Vector Machine (SVM), Decision Tree (DT), Random Forest (RF), K-Nearest Neighbor, Naive Bayes, gradient-boosted decision tree (GBDT) models (e.g., XGBoost, LightGBM, CatBoost), and Multi-layer Perceptron (MLP)~\cite{DelayNigam2017, DelayLi202302, DelayKhan2021, DelayTang2022, DelayVo2022, DelayHatipouglu2024, DelayAlfarhood2024}. Delay duration has also been modeled as regression, using Linear Regression, Lasso, Ridge, SVM, tree-based models, GBDTs, and MLP~\cite{DelayYu2019, DelayWu2019, DelayKhan2021, DelayWang2022, DelayReddy2023, DelayAlfarhood2024}.

As flight and weather features may be insufficient to characterize delay context, prior studies have incorporated additional tabular features such as traffic count~\cite{DelayShao2019}. Some works focused on departure delay prediction, introducing passenger-related features for delay distribution estimation~\cite{DelayBeltman2025}, or modeling IATA-coded delay causes via Adaptive Bidirectional Extreme Learning Machine (AB-ELM) classification~\cite{DelayKhan2024}. Other works incorporated sequences of tabular features, including prior delay information: \cite{DelayKim2016, DelayGui2020, DelayGopichand2024} employ long short-term memory (LSTM) models to classify delays, and Khanal et al.~\cite{Khanal2025} used a Gaussian Process to model delay duration from prior delay observations and flight dates. These frameworks require flight and weather data as fixed-size vectors, or rely on manual parsers to convert text-coded inputs such as Meteorological Aerodrome Report (METAR) data into numerical form. They are thus not readily applicable to variable-length Terminal Aerodrome Forecast (TAF) reports or Notices to Airmen (NOTAMs), which require semantic interpretation~\cite{notamevolve} and cannot be represented as fixed-size features.

Several recent works have applied spatiotemporal features for delay prediction. Models such as the LSTM-integrated RF model~\cite{DelayLi2022} and Convolutional Neural Network with LSTM (CNN-LSTM)~\cite{DelayLi202301} predict delays from weather, delay propagation, and traffic count. Qu et al.~\cite{DelayQu2023} modeled multi-flight delay propagation using attention-based architectures capturing flight chain dependencies across sequential aircraft operations, accounting for temporal coupling and propagation effects in aircraft rotations. Several recent studies have adopted graph neural networks: Cai et al.~\cite{DelayCai2022} employed a GCN-CNN architecture, where a Graph Convolutional Network (GCN) captures spatial dependencies in a time-evolving airport delay network and a CNN models daily and weekly delay patterns; \cite{DelayShen2024} proposed a hybrid framework combining diffusion GCNs with residual Gated Recurrent Units (GRUs) and federated learning across airport networks for privacy-preserving forecasting; and Li et al.~\cite{DelayLi2024} proposed FAST-CA, a graph-based spatio-temporal framework incorporating arrival-departure delay relationships and weather information via coupled attention to predict network-level delay propagation. However, spatiotemporal approaches often report relatively high prediction errors and limited goodness-of-fit, and graph-based inputs remain limited in capturing delay-related factors beyond inter-airport delay sequences and weather graphs, such as detailed airport-level airspace congestion. A recent trajectory-based approach~\cite{DelayChaudhuri2024} employs an attention-based LSTM combining trajectory, flight, and weather data to classify delays as early, on-time, or late, but considers only a single trajectory, which does not fully reflect airspace conditions.

Prior works are limited in two ways: tabular and sequential approaches rely on fixed-size features and manual parsers, making them inapplicable to text-based inputs such as TAF and NOTAMs; and spatiotemporal and graph-based approaches model delay propagation rather than underlying causes or airport-level airspace conditions, with a single trajectory further limited in capturing broader airspace conditions. We address these gaps jointly by modeling all information as tokens and incorporating multi-trajectory-based airspace conditions via cross-modality adaptation, enabling a more comprehensive context that we hypothesize will improve prediction accuracy.

\subsection{Cross-Modality Adaptation}\label{sec2.2}
Cross-modality adaptation is a technique for leveraging pretrained models, particularly LLMs, from one modality and adapting them to other modalities. Recent frameworks align time-series data with the language modality, enabling LLM-based time-series forecasting without requiring full fine-tuning of the pretrained LLM. LLMTIME \cite{llmtime} serializes continuous-valued time series into digit tokens, enabling direct use of pretrained LLMs and their native language modeling heads for forecasting. This technique leaves the LLM to infer temporal dependencies from digit-level semantics. Time-LLM \cite{timellm} projects feature patches and performs attention over word-prototypes to produce feature-wise patch embeddings, prepends a prompt prefix, and feeds them into a frozen LLM, followed by a learnable projection head to forecast future states. AutoTimes \cite{autotimes} partitions the time series into fixed-size feature-wise segments, projects them into embeddings, combines them with text-based positional encodings, and applies a learnable output projection after the frozen LLM. These works employ adaptation layers that encode feature-wise segment-level semantics for single-series forecasting from historical context alone, thereby forcing the LLM to infer inter-feature and inter-segment temporal relationships in order to reconstruct the instance-level semantics needed to understand maneuvering patterns. In contrast, we propose instance-level projection, a technique that adapts instance-level semantics learned by a pretrained encoder into the LLM embedding space. Such encoders, trained via representation learning techniques such as autoencoders~\cite{Olive2020DeepTrajectory} or contrastive learning \cite{TS2Vec, InfoTS, ATSCC}, capture feature dependencies and temporal dynamics in time-series data, such as aircraft trajectories, and produce instance-level representations that our adaptation layer maps directly into the LLM's embedding space.

\section{Methodology}\label{sec3}
\subsection{Problem Formulation}

\begin{figure*}[htbp]
    \centering
    \includegraphics[width=\textwidth]{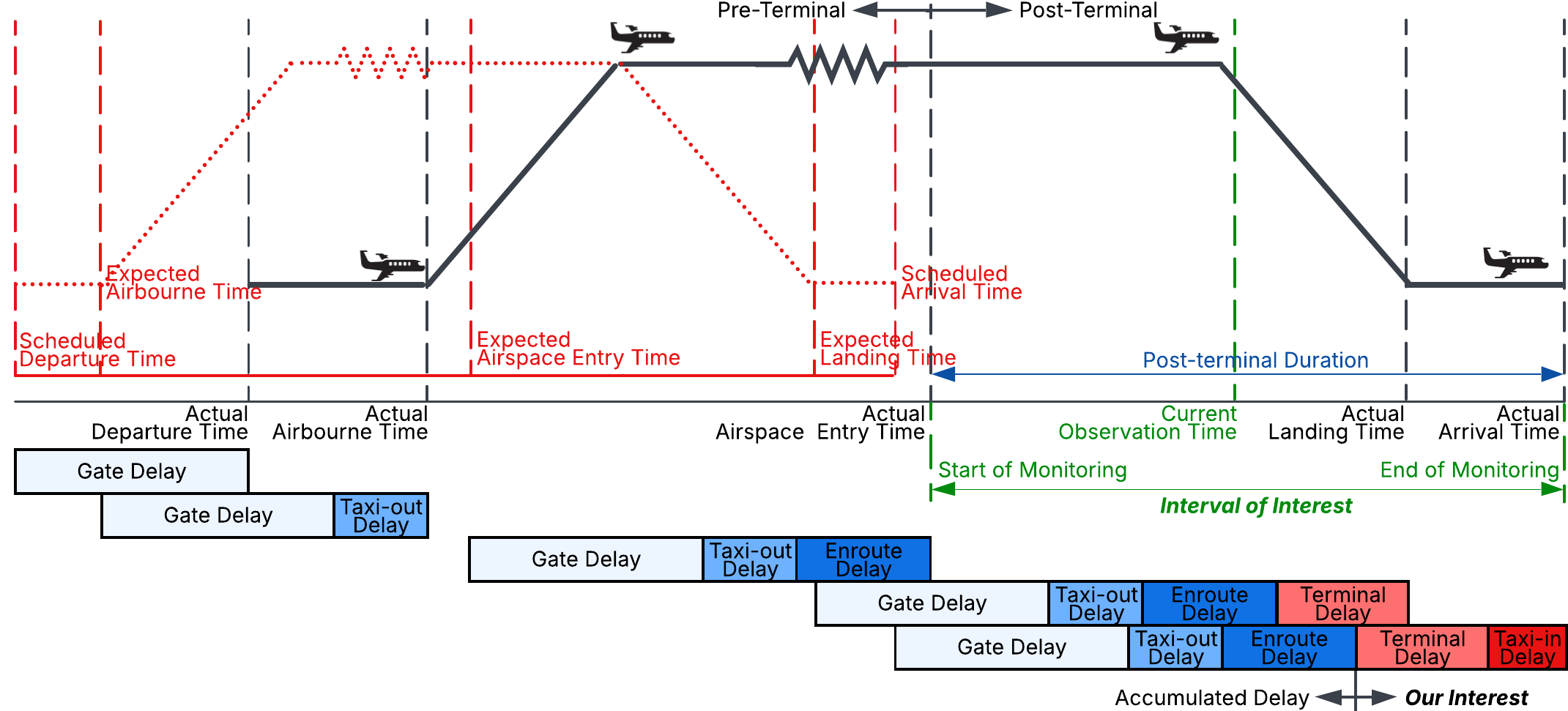}
    \caption{Flight Delay Accumulation of Aircraft Operations from Departure to Arrival}
    \label{fig:timehorizon}
\end{figure*}
Consider a flight with identifier $i$, whose operation is illustrated in Fig.~\ref{fig:timehorizon}. Key temporal markers defining a flight's timeline include scheduled/expected and actual departure times, airborne times, terminal maneuvering area (TMA) entry times, landing times, and arrival times. Total delay can be decomposed into phase-specific components representing the delay incurred at each stage of the flight~\cite{Mueller2002}. We define the pre-terminal phase as all operations prior to entering the TMA; accordingly, pre-terminal delay comprises gate, taxi-out, and en route delays.
\begin{align}
    D_{\text{pre},i} &= D_{\text{gate},i} + D_{\text{taxi-out},i} + D_{\text{enroute},i} \\
    D_{\text{pre},i} &= D_{\text{gate},i} + D_{\text{taxi-out},i} + (T_{\text{entry},i}^{\text{actual}} - T_{\text{entry},i}^{\text{expect}} - D_{\text{gate},i} - D_{\text{taxi-out},i} ) \\
    D_{\text{pre},i} &= T_{\text{entry},i}^{\text{actual}} - T_{\text{entry},i}^{\text{expect}}.
    \label{eq:delay_pre}
\end{align}

We define the post-terminal phase as the aircraft's operations after entering the TMA. The post-terminal delay includes terminal and taxi-in delays. These delay components can be rearranged as follows:
\begin{align}
    D_{\text{post},i} &= D_{\text{terminal},i} + D_{\text{taxi-in},i} \\
    D_{\text{post},i} &= (T_{\text{land},i}^{\text{actual}} - T_{\text{land},i}^{\text{expect}}) - (T_{\text{entry},i}^{\text{actual}} - T_{\text{entry},i}^{\text{expect}}) \nonumber \\ &\quad + (T_{\text{arrival},i}^{\text{actual}} - T_{\text{arrival},i}^{\text{schedule}}) - (T_{\text{land},i}^{\text{actual}} - T_{\text{land},i}^{\text{expect}}) \\
    D_{\text{post},i} &= (T_{\text{arrival},i}^{\text{actual}} - T_{\text{arrival},i}^{\text{schedule}}) - (T_{\text{entry},i}^{\text{actual}} - T_{\text{entry},i}^{\text{expect}}).
\end{align}

While prior works studied phase-specific departure delays~\cite{DelayBeltman2025, DelayKhan2024}, our formulation focuses on the post-terminal phase from the perspective of destination-airport ATC, tracking delay from TMA entry at \( T_{\text{entry},i}^{\text{actual}} \) until arrival at \( T_{\text{arrival},i}^{\text{actual}} \). Since \( T_{\text{arrival},i}^{\text{schedule}} \) is known and \( T_{\text{entry},i}^{\text{actual}} \) marks the start of monitoring, part of the delay can be inferred from these known values without explicit knowledge of gate, taxi-out, or en-route delays, as total delay can be rearranged as follows:
\begin{align}
    D_{\text{total},i} &= D_{\text{pre},i} + D_{\text{post},i} \\
    D_{\text{total},i} &= (T_{\text{entry},i}^{\text{actual}} - T_{\text{entry},i}^{\text{expect}}) + (T_{\text{arrival},i}^{\text{actual}} - T_{\text{arrival},i}^{\text{schedule}}) - (T_{\text{entry},i}^{\text{actual}} - T_{\text{entry},i}^{\text{expect}}) \\
    D_{\text{total},i} &= T_{\text{entry},i}^{\text{actual}} - T_{\text{arrival},i}^{\text{schedule}} + (T_{\text{arrival},i}^{\text{actual}} - T_{\text{entry},i}^{\text{actual}}) \\
    D_{\text{total},i} &= T_{\text{entry},i}^{\text{actual}} - T_{\text{arrival},i}^{\text{schedule}} + \Delta t_{\text{post},i}.
\end{align}

The true post-terminal duration, denoted \( \Delta t_{\text{post}, i} \), remains unknown for \( t \in [T_{\text{entry},i}^{\text{actual}}, T_{\text{arrival},i}^{\text{actual}}) \) since \( T_{\text{arrival},i}^{\text{actual}} \) has not yet been observed; accordingly, we aim to estimate \( \Delta t_{\text{post}, i} \) using available contextual information. Unlike prior spatiotemporal or temporal models, prior delay information and pre-entry delays are absorbed into the known term \(T_{\text{entry},i}^{\text{actual}} - T_{\text{arrival},i}^{\text{schedule}}\) in Eq.~\ref{eq:delay_pre}, leaving only \( \Delta t_{\text{post},i} \) to be predicted. Thus, the estimated total delay is given by:
\begin{equation}
    \hat{D}_{\text{total},i} = T_{\text{entry},i}^{\text{actual}} - T_{\text{arrival},i}^{\text{schedule}} + \Delta \hat{t}_{\text{post},i}.
\end{equation}

Estimating \(\Delta t_{\text{post},i}\) is advantageous because it often aligns with maneuvering procedures, where similar procedures yield similar durations, though variations still arise from many factors. We hypothesize that accurate estimation requires integrating flight information, weather conditions, non-periodic operational events, and traffic congestion. We thus define a scenario \(s_{i,t}\) of flight $i$ at time $t$ as a set of textual prompts encoding flight, METAR, TAF, NOTAM, and trajectory-based airspace information:
\begin{equation}
\begin{aligned}
s_{i,t} &= \{ P^{\text{F}}_{i,t},P^{\text{M}}_{t},P^{\text{T}}_{t},P^{\text{N}}_{t},X^f_{i,t},X^a_{i,t},X^p_{i,t} \}, \\
P^{\text{F}}_{i,t},P^{\text{M}}_{t},P^{\text{T}}_{t},P^{\text{N}}_{t} &\in \mathbb{Z}^{L^{j}_{i,t}}, \quad j \in \{\text{F}, \text{M}, \text{T}, \text{N}\}, \\
X^f_{i,t},X^a_{i,t},X^p_{i,t} &\in \mathbb{R}^{N^{k}_{i,t} \times \mathcal{T}^{k}_{i,t} \times 9}, \quad k \in \{f, a, p\}.
\end{aligned}
\label{eq:scenario}
\end{equation}
The prompts include flight information (\(P^{\text{F}}_{i,t}\)), weather conditions via METAR (\(P^{\text{M}}_{t}\)) and TAF (\(P^{\text{T}}_{t}\)), and NOTAMs (\(P^{\text{N}}_{t}\)). This paper introduces a trajectory-based representation of airspace conditions that effectively captures congestion rather than the prior traffic-count feature used in \cite{DelayShao2019, DelayLi2022, DelayLi202301}. A trajectory is represented as a time series of aircraft positions. To characterize the airspace, we consider three trajectory groups: the focusing trajectory \(X^{f}_{i,t}\), the active trajectories \(X^{a}_{i,t}\) of aircraft operating in the TMA at time \(t\), and the completed prior trajectories \(X^{p}_{i,t}\), which provide additional context. The model is formulated as \( \hat{y}_i = f_\theta(s_{i,t}) \), where \( \hat{y}_i \equiv \Delta\hat{t}_{\text{post}, i} \), the two notations used interchangeably. \(f_\theta(\cdot)\) is trained as a regression task, minimizing the error between predicted post-terminal duration \(\hat{y}_i\) and ground-truth \(y_i\).

\subsection{Dataset Preparation}
\subsubsection{Flight Information Features}
The retrieved 2022 arrival records for Incheon Airport from Airportal~\cite{MLIT2024Airportal} include scheduled arrival times, airline names, flight identifiers, and departure/destination airport codes and names. We matched airports to their geodetic coordinates and altitude, computed the great-circle distance between departure and destination, and categorized flight haul type (short-, medium-, or long-haul) accordingly, following Wragg's aviation dictionary~\cite{Wragg1973}. Arrival date and day of week are defined in Korea Standard Time (KST). Certain fields were synthesized to complete the dataset: aircraft type and registration were inferred from the most frequently used aircraft model on each route, and wake turbulence category was assigned based on aircraft type per ICAO Doc 4444~\cite{ICAO4444}. For each flight \( i \), the flight information features \( \mathbf{F}^{\text{F}}_i \) (Table~\ref{tab:flight_data}), together with \( T_{\text{entry},i}^{\text{actual}} \) and the current time \( t \), were used to construct the prompt \( P^{\text{F}}_{i,t} \), randomly selected from ten predefined formats (see supplemental code) to introduce variation in prompt formatting.

\begin{table*}[ht]
\centering
\caption{Flight Information Feature Descriptions}
\begingroup
\renewcommand{\arraystretch}{0.85}
\resizebox{0.6\textwidth}{!}{
\begin{tabular}{lll}
\toprule
\textbf{Flight Information} & \textbf{Description} & \textbf{Type} \\
\midrule
\texttt{airline\_name\_english} & Airline name in English & \texttt{string} \\
\texttt{callsign\_code\_iata} & IATA-format callsign & \texttt{string} \\
\texttt{callsign\_code\_icao} & ICAO-format callsign & \texttt{string} \\
\texttt{haul} & Flight haul type & \texttt{string} \\
\midrule
\texttt{dep\_code\_iata} & Departure airport IATA code & \texttt{string} \\
\texttt{dep\_code\_icao} & Departure airport ICAO code & \texttt{string} \\
\texttt{dep\_name\_english} & Departure airport name & \texttt{string} \\
\texttt{dep\_lat} & Departure airport latitude & \texttt{float} \\
\texttt{dep\_lon} & Departure airport longitude & \texttt{float} \\
\texttt{dep\_altitude} & Departure airport altitude (ft) & \texttt{float} \\
\midrule
\texttt{dest\_code\_iata} & Destination airport IATA code & \texttt{string} \\
\texttt{dest\_code\_icao} & Destination airport ICAO code & \texttt{string} \\
\texttt{dest\_name\_english} & Destination airport name & \texttt{string} \\
\texttt{dest\_lat} & Destination airport latitude & \texttt{float} \\
\texttt{dest\_lon} & Destination airport longitude & \texttt{float} \\
\texttt{dest\_altitude} & Destination airport altitude (ft) & \texttt{float} \\
\midrule
\texttt{distance} & Great-circle distance (km) & \texttt{float} \\
\texttt{actual\_entry\_time} & Time of first ADS-B appearance (UTC) & \texttt{datetime} \\
\texttt{sched\_time\_utc} & Scheduled arrival time (UTC) & \texttt{datetime} \\
\texttt{date} & Scheduled flight date of arrival (KST) & \texttt{datetime} \\
\texttt{day\_of\_week} & Scheduled day of week of arrival (KST) & \texttt{string} \\
\midrule
\texttt{aircraft\_type} & Aircraft type & \texttt{string} \\
\texttt{aircraft\_registration} & Aircraft registration code & \texttt{string} \\
\texttt{wake\_turbulence\_cat} & Wake turbulence category & \texttt{string} \\
\bottomrule
\end{tabular}
}
\endgroup
\label{tab:flight_data}
\end{table*}

\subsubsection{Textual Data}
We retained the original aeronautical-coded forms of METARs, TAFs, and NOTAMs to highlight a design aspect in which the framework does not require rigid rule-based parsing. METARs and TAFs were mass-downloaded from Ogimet~\cite{Ogimet}, while NOTAMs were collected from AIM Korea~\cite{AIMKorea}. The METAR was selected as the most recent report released before time \(t\), within the standard 30-minute update interval. TAFs and NOTAMs explicitly define their active periods: TAFs are typically reported every six hours, and since multiple forecasts may overlap at a given time, we selected the most recent valid TAF at time \(t\); for NOTAMs, multiple notices can be active concurrently, so we collected all active notices. The METAR, TAF, and NOTAMs queried at time \(t\) were directly used to construct the prompts \(P^{\text{M}}_t\), \(P^{\text{T}}_t\), and \(P^{\text{N}}_t\), with the prefixes ``METAR in effect:'', ``TAF in effect:'', and ``Active NOTAMs:'', respectively.

\subsubsection{Trajectory Data}
\begin{figure}[htbp]
    \centering
    \includegraphics[width=0.55\textwidth]{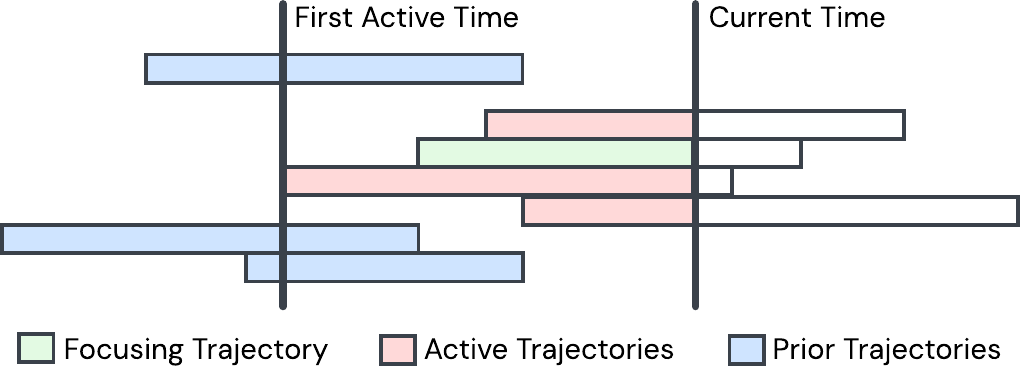}
    \caption{Temporal Relationship Between Focusing, Active, and Prior Trajectories}
    \label{fig:trajtime}
\end{figure}

Automatic Dependent Surveillance-Broadcast (ADS-B) data were sourced from the OpenSky database~\cite{Schafer2014OpenSky}. ADS-B data for arrival and departure flights were queried using flight identification numbers from Airportal~\cite{MLIT2024Airportal} and filtered within latitude 36.6-37.9 and longitude 125.1-127.5 to cover the Incheon Airport TMA defined in the area chart. We extracted positional time series from the ADS-B data, including geodetic latitude, longitude, and altitude, and transformed them to the airport-centered ENU (East-North-Up) frame. Trajectories were resampled to 5-second intervals without interpolating to a fixed number of timestamps. Each trajectory was scaled by dividing by 120 kilometers, following ATFMTraj preprocessing~\cite{atfmtraj}. Unlike intensively preprocessed pretraining data, we avoided trajectory smoothing to preserve real-world characteristics. A complete trajectory of flight \(i\) consists of states \( x_{i,\kappa} = \{x_{i,\kappa}^x, x_{i,\kappa}^y, x_{i,\kappa}^z\} \) over all timestamps. To enhance expressiveness, additional geometric features are computed for each timestep \( \kappa \) in a time-series instance as:
\begin{align}
    \{x_{i,\kappa}^{u_x}, x_{i,\kappa}^{u_y}, x_{i,\kappa}^{u_z}\} &= \frac{x_{i, \kappa+1} - x_{i,\kappa}}{\left\| x_{i, \kappa+1} - x_{i,\kappa} \right\|}, \label{eq:direction_vector} \\
    x_{i,\kappa}^{r} &= \sqrt{\left(x_{i,\kappa}^x\right)^2 + \left(x_{i,\kappa}^y\right)^2}, \\
    x_{i,\kappa}^{\sin\theta} &= \sin\left(\arctan2(x_{i,\kappa}^y, x_{i,\kappa}^x)\right), \\
    x_{i,\kappa}^{\cos\theta} &= \cos\left(\arctan2(x_{i,\kappa}^y, x_{i,\kappa}^x)\right).
    \label{eq:polar_coordinates}
\end{align}
These features are collected at each timestep $\kappa$ into the feature set:
\begin{equation}
    x_{i,\kappa} = \{x_{i,\kappa}^{x}, x_{i,\kappa}^{y}, x_{i,\kappa}^{z}, x_{i,\kappa}^{u_x}, x_{i,\kappa}^{u_y}, x_{i,\kappa}^{u_z}, x_{i,\kappa}^{r}, x_{i,\kappa}^{\sin\theta}, x_{i,\kappa}^{\cos\theta}\}.
\label{eq:trajfeature}
\end{equation}
In $s_{i,t}$, there are three types of trajectories: focusing, active, and prior. Their temporal relationships are illustrated in Fig.~\ref{fig:trajtime}. We use the notation $i,t/j$ to denote flight $j$'s trajectory considered relative to anchor flight $i$ at time $t$. Definitions are as follows:
\begin{itemize}
    \item \textbf{Focusing trajectory}: The trajectory of flight \(i\) comprises the focusing aircraft states over the interval \([T_{\text{entry}, i}^{\text{actual}},\, t]\). We define \( X^{f}_{i,t} \in \mathbb{R}^{1 \times \mathcal{T}^f_{i,t} \times 9} \), where \( \mathcal{T}^f_{i,t} \) denotes the number of observed aircraft trajectory states.
    
    \item \textbf{Active trajectories}: The trajectories of other aircraft in the TMA at time \(t\), excluding flight \(i\). A trajectory of flight \(j\) is considered active at time \(t\) if its state \(x_{j,t} \in \mathbb{R}^9\) exists, with \(\mathcal{T}^{a}_{i,t/j}\) observations over the interval \([T_{\text{entry},j}^{\text{actual}},\, t]\) or \([T_{\text{departure},j}^{\text{actual}},\, t]\). Collecting all such trajectories \(X^{a}_{i,t/j}\) into \(X^{a}_{i,t}\), we nan-pad and define \( X^{a}_{i,t} \in \mathbb{R}^{N^a_{i,t} \times \mathcal{T}^{a}_{i,t} \times 9} \), where \( N^a_{i,t} \) is the number of active trajectories and \( \mathcal{T}^{a}_{i,t} = \max_{X^{a}_{i,t/j}\in X^{a}_{i,t}}(\mathcal{T}^a_{i,t/j})\).
    
    \item \textbf{Prior trajectories}: Using \(\min_{X_{i,t/j} \in X^{a}_{i,t} \cup X^{f}_{i,t}} (T_{\text{entry},j}^{\text{actual}})\) as reference, we collect all completed prior trajectories \(X^{p}_{i,t/j}\) whose state \(x_{j,\kappa} \in \mathbb{R}^9\) exists at this reference time, each consisting of \(\mathcal{T}^{p}_{i,t/j}\) observations over the interval \([T_{\text{entry},j}^{\text{actual}},\, T_{\text{arrival},j}^{\text{actual}}]\) or \([T_{\text{departure},j}^{\text{actual}},\, T_{\text{exit},j}^{\text{actual}}]\). As with active trajectories, we nan-pad and define \( X^{p}_{i,t} \in \mathbb{R}^{N^p_{i,t} \times \mathcal{T}^{p}_{i,t} \times 9} \), where \( N^p_{i,t} \) denotes the number of prior trajectories and \( \mathcal{T}^{p}_{i,t} = \max_{X^{p}_{i,t/j}\in X^{p}_{i,t}}(\mathcal{T}^p_{i,t/j})\).
\end{itemize}

The trajectories \( X^f_{i,t} \), \( X^a_{i,t} \), and \( X^p_{i,t} \) represent the query at time \( t \) for flight \( i \) under scenario \( s_{i,t} \). Intuitively, these trajectory groups provide a comprehensive semantic view of the airspace, similar to the situational awareness of human ATCs: \( X^f_{i,t} \) captures the target aircraft's movement, \( X^a_{i,t} \) reflects current congestion and airspace conditions, and \( X^p_{i,t} \) encodes historical patterns that may influence ongoing operations and implicitly reflect ground traffic, enabling the model to directly estimate \(\Delta t_{\text{post},i}\) from trajectory information.

\subsubsection{Dataset Annotation}
For supervised training, we annotated the regression label \(y_i\) corresponding to \(s_{i,t}\) as \(\Delta t_{\text{post},i}\), computed from \(T_{\text{entry},i}^{\text{actual}}\) and \(T_{\text{arrival},i}^{\text{actual}}\). \(T_{\text{entry},i}^{\text{actual}}\) is defined as the first ADS-B broadcast timestamp after the aircraft enters Incheon Airport's TMA, based on OpenSky data~\cite{Schafer2014OpenSky} and following the ICAO Global Air Navigation Plan (GANP) Doc 9750~\cite{ICAO9750}, while \(T_{\text{arrival},i}^{\text{actual}}\) is obtained from official Airportal records~\cite{MLIT2024Airportal}. The ground-truth label is:
\begin{equation}
    y_i \equiv \Delta t_{\text{post},i} = T_{\text{arrival},i}^{\text{actual}} - T_{\text{entry},i}^{\text{actual}}.
\end{equation}
We repeated this procedure for all flights \(i\) within each month, sampling two time points \(t\) per flight, since any \(t \in [T_{\text{entry},i}^{\text{actual}},\, T_{\text{arrival},i}^{\text{actual}}]\) yields the same \(y_i\), resulting in 12 monthly datasets (\(\mathcal{S}\)) for 2022. For each dataset, instances are sorted by \(t\): the most recent 10\% form the test set, the next 10\% form the validation set, and the remainder form the training set. All monthly datasets are publicly available in the supplementary Hugging Face repository accompanying this paper.

\subsection{Neural Network Architecture}
\begin{figure*}[htbp]
    \centering
    \includegraphics[width=\textwidth]{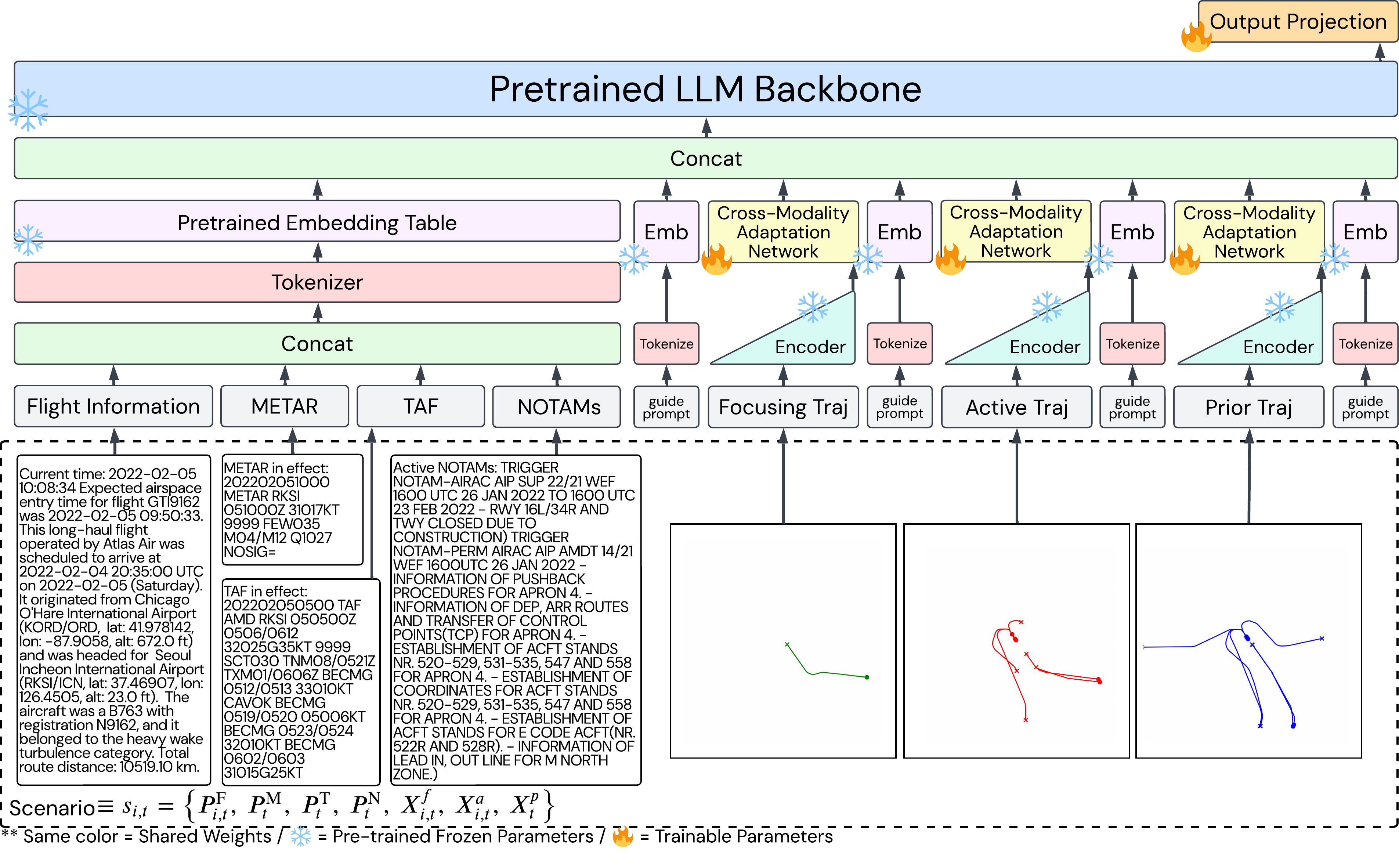}
    \caption{Overall Architecture of the LLM4Delay Framework}
    \label{fig:LLMxDelay}
\end{figure*}

\subsubsection{Natural Language Prompt Embedding}
To process \(P^{\text{F}}_{i,t}\), \(P^{\text{M}}_{t}\), \(P^{\text{T}}_{t}\), and \(P^{\text{N}}_{t}\), we use the tokenizer \( \text{T}(\cdot) \) and embedding table \(\text{Emb}[\cdot]\) provided with the selected LLM. \( \text{T}(\cdot) \) transforms each combined prompt \( P_{i,t} = P^{\text{F}}_{i,t} \, \| \, P^{\text{M}}_t \, \| \, P^{\text{T}}_t \, \| \, P^{\text{N}}_t\) into the tokenized sequence \(\text{T}(P_{i,t}) \in \mathbb{N}^{L_{i,t}}\), where \( L_{i,t} \) denotes the token sequence length. Each subword token is then mapped to a continuous embedding using \( \text{Emb} \in \mathbb{R}^{V \times d} \), where \( V \) is the vocabulary size and \( d \) is the LLM's embedding dimension:
\begin{equation}
    \mathbf{Z}^p_{i,t} = \text{Emb}[\text{T}(P_{i,t})],
    \label{eq:tokenize}
\end{equation}
where \( \mathbf{Z}^p_{i,t} \in \mathbb{R}^{L_{i,t} \times d} \) is the sequence of token embeddings for \(P_{i,t}\). \(\text{Emb}[\cdot]\) is kept fixed during training to preserve pretrained language understanding and ensure consistent alignment with the LLM backbone.

\subsubsection{Cross-Modality Adaptation of Trajectory Data}
\begin{figure*}[htbp]
    \centering
    \includegraphics[width=\textwidth]{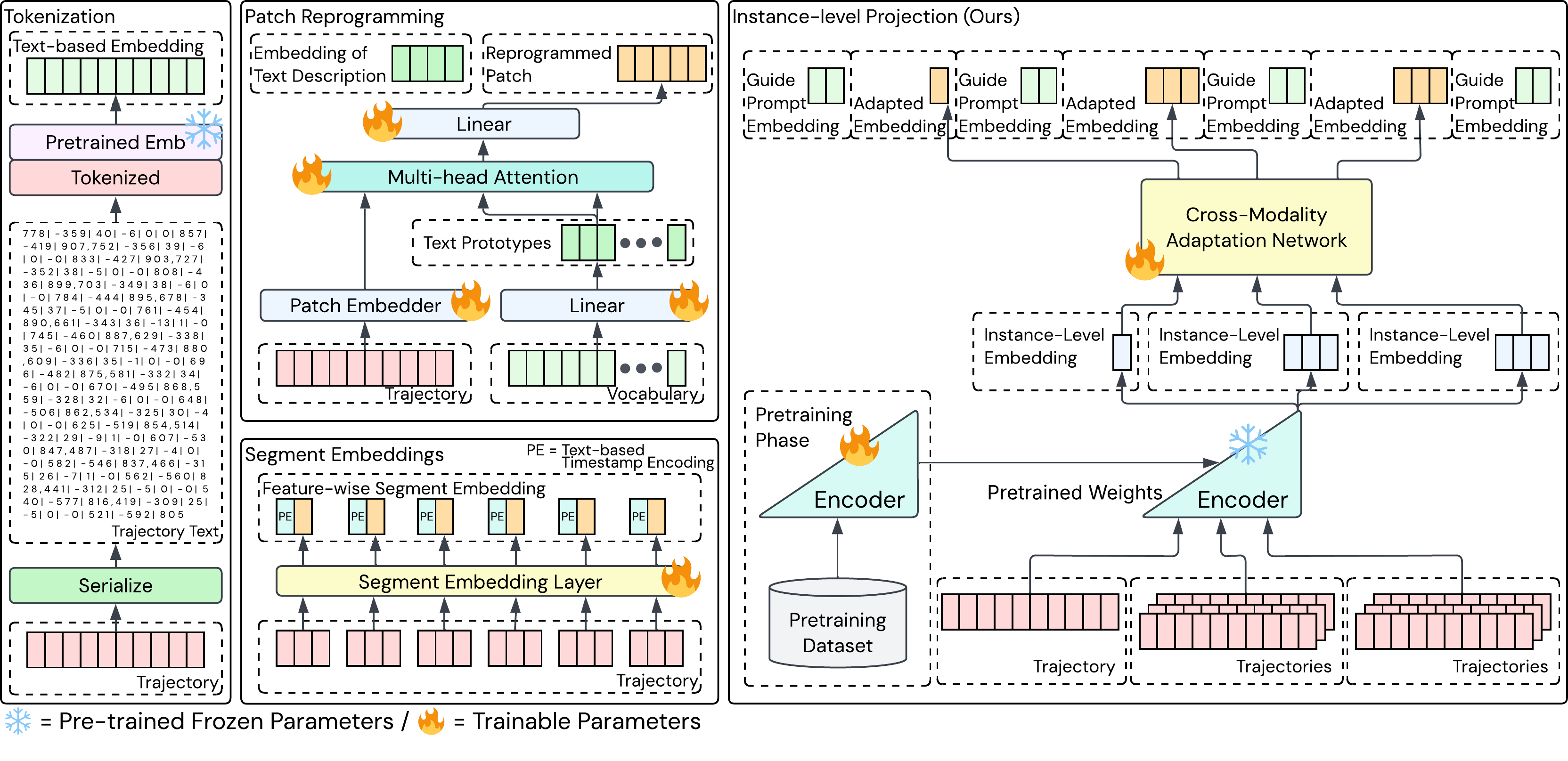}
    \caption{Comparison of Different Time-Series-to-Language Cross-Modality Adaptation Techniques}
    \label{fig:fusion}
\end{figure*}

As discussed in Section~\ref{sec2.2}, recent cross-modality adaptation works (Fig.~\ref{fig:fusion}) encode digit-level or feature-wise segment-level semantics rather than instance-level semantics, leaving the LLM to infer inter-feature and inter-segment temporal relationships on its own. However, \( \Delta t_{\text{post},i} \) depends on trajectory patterns reflecting the type of maneuvering procedure, underscoring the importance of instance-level semantics. Accordingly, our proposed technique, instance-level projection, employs a frozen pretrained trajectory encoder that has already learned inter-feature correlations and instance-level semantics. Moreover, since \( \Delta t_{\text{post},i} \) depends not only on a trajectory's own history but also on surrounding air traffic, our method extracts instance-level semantics from multiple trajectories across three trajectory groups and aligns their representations with the LLM embedding space, directly mapping collective trajectory semantics to \( \Delta t_{\text{post},i} \).

Prior to training $f_\theta(\cdot)$, we pretrain an encoder $f_{\text{enc}}(\cdot)$ via self-supervised learning on the ATFMTraj dataset~\cite{atfmtraj}, using arrival and departure trajectories from Incheon Airport, with feature extraction and 5-second downsampling identical to our dataset preparation. Candidate representation learning methods include a Temporal Convolutional Network AE (TCN-AE) and contrastive approaches (TS2Vec~\cite{TS2Vec}, InfoTS~\cite{InfoTS}, and ATSCC~\cite{ATSCC}). The pretrained encoder $f_{\text{enc}}(\cdot)$ extracts trajectory representations and remains frozen during training: $X^f_{i,t}$, $X^a_{i,t}$, and $X^p_{i,t}$ are compressed into instance-level embeddings by $f_{\text{enc}}(\cdot)$, given by:
\begin{equation}
    Z^{k}_{i,t} = f_{\text{enc}}(X^{k}_{i,t}),
    \label{eq:fatscc}
\end{equation}
where \(Z^{k}_{i,t} \in \mathbb{R}^{N^{k}_{i,t} \times 320}\) and \(k \in \{f, a, p\}\), representing \(N^{k}_{i,t}\) instance-level embeddings of dimension 320, following reproduction details in \cite{TS2Vec, InfoTS, ATSCC}. As shown in Fig.~\ref{fig:LLMxDelay}, $f_{\text{enc}}(\cdot)$ is placed at the start of the pipeline, allowing trajectories to be pre-encoded into instance-level representations \(Z^{k}_{i,t}\), which are then subjected to dropout with a rate of 0.3. This design improves memory efficiency by avoiding nested sequence encoding while retaining trajectory information. We then employ a lightweight MLP-based cross-modality adaptation network \(f_{\text{xa}}(\cdot)\) that projects instance-level embeddings \(Z^{k}_{i,t}\) into the LLM embedding space, given by:
\begin{equation}
    \mathbf{Z}^{k}_{i,t} = f_{\text{xa}}(Z^{k}_{i,t}),
    \label{eq:fxa}
\end{equation}
where \( \mathbf{Z}^{k}_{i,t} \in \mathbb{R}^{N^{k}_{i,t} \times d} \) is the embedding of trajectory group \( k \in \{f, a, p\} \). \(f_{\text{xa}}(\cdot)\) consists of two linear layers with Gaussian Error Linear Unit (GELU) activations and a dropout rate of 0.35 applied after each activation: the first layer maps the 320-dimensional embedding to dimension $d$, while the second preserves it. This lightweight design introduces minimal computational overhead while bridging the trajectory encoder and LLM. We then insert trajectory embeddings \(\mathbf{Z}^{k}_{i,t}\) at designated positions between guiding prompts to distinguish trajectory groups, yielding the trajectory-informed sequence: 
\begin{align}
    \mathbf{Z}^{T}_{i,t} = \text{Concat}\big[&\text{Emb}[\text{T}(P_{st,1})], \nonumber\\
    &\ \mathbf{Z}^{f}_{i,t},\ \text{Emb}[\text{T}(P_{st,2})],\nonumber\\
    &\ \mathbf{Z}^{a}_{i,t},\ \text{Emb}[\text{T}(P_{st,3})],\nonumber\\
    &\ \mathbf{Z}^{p}_{i,t},\ \text{Emb}[\text{T}(P_{st,4})]\big],
    \label{eq:trajinform}
\end{align}
where \(\mathbf{Z}^{T}_{i,t} \in \mathbb{R}^{(L_{st,1}+1+L_{st,2}+N^{a}_{i,t}+L_{st,3}+N^{p}_{i,t}+L_{st,4}) \times d}\), and \(L_{st,k}\) is the length of the \(k\)-th guiding prompt.

\subsubsection{Large Language Model Backbone}
The pretrained LLM backbone \(f_{\text{llm}}(\cdot)\) is repurposed, with its language modeling head replaced by a regression head to predict \(\Delta t_{\text{post},i}\). We consider open LLMs with one billion parameters or fewer, as the task involves scalar-value regression and does not require large generative capacity; these LLMs also run efficiently on a single consumer-grade GPU, supporting broader deployment. Candidate LLMs include LLaMA3.2-1B, LLaMA3.2-1B-Instruct, Qwen3-0.6B, Qwen3-0.6B-Base, and Pythia-1B. All \(f_{\text{llm}}(\cdot)\) parameters are kept frozen to retain pretrained linguistic capabilities while maintaining memory efficiency. \(f_{\text{llm}}(\cdot)\) processes multimodal inputs by concatenating \(\mathbf{Z}^{p}_{i,t}\) and \(\mathbf{Z}^{T}_{i,t}\):
\begin{equation}
    \mathbf{Z}_{i,t} = \text{Concat}\big[\mathbf{Z}^{p}_{i,t}, \mathbf{Z}^{T}_{i,t}\big].
\end{equation}
The sequence \( \mathbf{Z}_{i,t} \) is then passed through the LLM backbone:
\begin{equation}
    \mathbf{h}_{i,t} = f_{\text{llm}}(\mathbf{Z}_{i,t})[-1].
\end{equation}
Due to the autoregressive nature of causal LLMs, we extract only the last hidden state \(\mathbf{h}_{i,t} \in \mathbb{R}^{d}\) as the summary of \(\mathbf{Z}_{i,t}\), used by the output regression head.

\subsubsection{Output Regression Head}
An MLP head \( f_h(\cdot) \) is attached to the backbone to map the multimodal hidden state \(\mathbf{h}_{i,t} \in \mathbb{R}^{d}\), which encodes scenario \(s_{i,t}\), into a scalar prediction. The regression head comprises three linear layers: the first maintains the LLM hidden size, the second halves it, and the third outputs a scalar prediction, with GELU activation and a 0.3 dropout rate following each of the first two layers. The model output is:
\begin{equation}
    \hat{y}_i = f_h(\mathbf{h}_{i,t}).
\end{equation}
This regression head repurposes pretrained LLMs for the regression task, producing \(\hat{y}_i\) from the rich multimodal context encoded in \( \mathbf{h}_{i,t} \).

\subsection{Training Pipeline}
With the encoder \(f_{\text{enc}}(\cdot)\) frozen, we pre-encode and cache trajectory embeddings \(Z^{k}_{i,t}\) for all \(s_{i,t} \in \mathcal{S}\) and \(k \in \{f,a,p\}\), reducing computational overhead during training. During the training loop, for each batch $\mathcal B=\{(s_{i,t}, y_i)\}_{i=1}^B$, \(P_{i,t}\) is tokenized and embedded to form the prompt embedding \( \mathbf{Z}^p_{i,t} \). Separately, \(Z^{k}_{i,t}\) is projected through the trainable \(f_{\text{xa}}(\cdot)\) and combined with embedded static guide prompts to form the trajectory-informed sequence \(\mathbf{Z}^T_{i,t}\). These are concatenated into \(\mathbf{Z}_{i,t}\) and processed by the frozen \(f_{\text{llm}}(\cdot)\) to produce the final hidden state \(\mathbf{h}_{i,t}\), which is fed to the trainable \(f_h(\cdot)\) to output \(\hat{y}_i \equiv \Delta\hat{t}_{\text{post},i}\). The ground-truth \( y_i \equiv \Delta t_{\text{post},i}\) is standardized prior to training. \(f_\theta(\cdot)\) is trained using the Smooth L1 Loss:
\begin{equation}
\mathcal{L}_{\text{smoothL1}}(\hat{y}, y) = 
\begin{cases}
0.5 \cdot (\hat{y} - y)^2, & \text{if } |\hat{y} - y| < 1 \\
|\hat{y} - y| - 0.5, & \text{otherwise}
\end{cases},
\label{eq:loss}
\end{equation}
which is robust to outliers and in some cases prevents exploding gradients~\cite{fastrcnn}. Training was performed for 15 epochs using the AdamW optimizer with a learning rate of \(1\times10^{-5}\), weight decay of \(1\times10^{-5}\), and a batch size of 4. Pretrained components \(\text{Emb}[\cdot]\), \(f_{\text{llm}}(\cdot)\), and \(f_{\text{enc}}(\cdot)\) are frozen during training, avoiding intensive gradient computation and memory allocation; only \( f_{\text{xa}}(\cdot) \) and \( f_h(\cdot) \) are updated (Fig.~\ref{fig:LLMxDelay}). All training and evaluation were conducted using Python 3.11.8 and PyTorch 2.0.1 with CUDA Toolkit 11.8, running on an NVIDIA GeForce RTX 4090 GPU.

\algrenewcommand\textproc{} 
\algrenewcommand\algorithmicindent{1.2em}

\begin{algorithm}
\small
\caption{Training Pipeline of LLM4Delay}
\label{alg:training}
\begin{algorithmic}[1]
\State \textbf{Input:} Scenario dataset $\mathcal{S} = \{(s_{i,t}, y_i)\}_{i=1}^N$,  where \\ \(s_{i,t} = \{ P^{\text{F}}_{i,t}, P^{\text{M}}_{t}, P^{\text{T}}_{t}, P^{\text{N}}_{t}, X^f_{i,t}, X^a_{i,t}, X^p_{i,t}\}\)
\State \textbf{Output:} Trained model parameter $\theta$
\For {$s_{i,t}$ in $\mathcal S$}
  \State $Z^{k}_{i,t} = f_{\text{enc}}(X^{k}_{i,t}), \quad k \in \{f, a, p\}$
\EndFor

\For{each epoch}
    \For{each batch $\mathcal B\subset\mathcal S$}
        \For{each scenario $(s_{i,t}, y_i)$ in $\mathcal{B}$}
        \State Build prompt $P_{i,t}=P^{\text{F}}_{i,t} \, \| \, P^{\text{M}}_t \, \| \, P^{\text{T}}_t \, \| \, P^{\text{N}}_t$
        \State $\mathbf Z^p_{i,t}=\mathrm{Emb}[\text{T}(P_{i,t})]$
        \State $\mathbf Z^{k}_{i,t}=f_{\text{xa}}(Z^{k}_{i,t}), \quad k \in \{f, a, p\}$
        \State \vspace{-4pt}
                {\setlength{\jot}{0.0pt}
                \begin{math}
                \begin{aligned}
                \mathbf{Z}^{T}_{i,t} = \text{Concat}\big[&\text{Emb}[\text{T}(P_{st,1})], \\
                    &\mathbf{Z}^{f}_{i,t},\ \text{Emb}[\text{T}(P_{st,2})], \\
                    &\mathbf{Z}^{a}_{i,t},\ \text{Emb}[\text{T}(P_{st,3})], \\
                    &\mathbf{Z}^{p}_{i,t},\ \text{Emb}[\text{T}(P_{st,4})]\big]
                \end{aligned}
                \end{math}}
                \vspace{4pt}
        \State $\mathbf{Z}_{i,t} = \text{Concat}\big[\mathbf{Z}^{p}_{i,t}, \mathbf{Z}^{T}_{i,t}\big] $
        \State $\mathbf{h}_{i,t} = f_{\text{llm}}(\mathbf{Z}_{i,t})[-1]$
        \State $\hat y_i=f_h(\mathbf h_{i,t})$
        \EndFor
    \State $\mathcal L=\frac{1}{|\mathcal B|}\sum_i\mathcal{L}_{\text{smoothL1}}(\hat y_i,y_i)$
    \State Backpropagate and update parameters in $f_{\text{xa}}(\cdot)$ and $f_{h}(\cdot)$
    \EndFor
\EndFor
\end{algorithmic}
\end{algorithm}

\section{Results and Discussion}\label{sec4}
\subsection{Comparison Study with Delay Prediction Baselines}
This section evaluates LLM4Delay against baseline tabular- and trajectory-based delay prediction models from existing ATM studies. As our task targets single-airport, flight-wise delay estimation, spatiotemporal (multi-airport, network-level) and temporal delay models were excluded; the latter is unaligned with our scope since our formulation (Fig.~\ref{fig:timehorizon}) already captures past and pre-entry information via $(T_{\text{entry},i}^{\text{actual}} - T_{\text{arrival},i}^{\text{schedule}})$. We thus compared against baselines matching our problem setting and data structure. Experiments were conducted on 12 monthly datasets, training on each month's split and evaluating at the checkpoint with the lowest validation loss. For each baseline, we provided the most comprehensive set of inputs the method supports, subject to its modeling constraints. Performance is reported over the 12 monthly datasets using Mean Absolute Error (MAE), Mean Squared Error (MSE), and the coefficient of determination ($R^2$) for $\Delta t_{\text{post}, i}$.

Tabular-based models have been adopted in prior works~\cite{DelayNigam2017, DelayLi202302, DelayKhan2021, DelayTang2022, DelayVo2022, DelayHatipouglu2024, DelayAlfarhood2024, DelayYu2019, DelayWu2019, DelayWang2022, DelayReddy2023} due to their compatibility with tabular flight and current weather information. The tabular flight information $F^{\text{F}}_{i}$ was derived from $\textbf{F}^{\text{F}}_{i}$ in Table~\ref{tab:flight_data} by converting strings into categorical features and removing unnecessary attributes, such as aircraft registration, IATA-coded information, departure airport names where ICAO codes suffice, and destination airport details, since all flights arrive at Incheon. We appended the traffic count feature \(N^a_{i,t}\) to help the model capture airspace congestion, as in \cite{DelayShao2019, DelayLi2022, DelayLi202301}, and extracted current weather features from METAR, parsed into tabular form \(F^{\text{M}}_t\) with an open-source METAR parser~\cite{pythonmetar}. Five tabular methods are considered: Linear Regression, SVM, Random Forest, and XGBoost implemented in scikit-learn, and a four-hidden-layer MLP with 1024 units per layer, ReLU activations, and a dropout rate of 0.1.

For trajectory-based methods, prior work~\cite{DelayChaudhuri2024} used only the focusing trajectory. Following this setting, we retained the tabular features and incorporated \(X^f_{i,t}\) as the time-series trajectory input. All trajectory-based models consist of a tabular feature extractor, a trajectory encoder, and a regression bottleneck. The tabular feature extractor is a two-layer MLP with 512 units per layer, ReLU activations, and a 0.1 dropout rate. The trajectory encoder varies across the following architectures:
\begin{itemize}
    \item \textbf{LSTM}: An LSTM encoder following the architecture described in \cite{DelayChaudhuri2024}, where the final hidden state serves as the sequence representation.
    \item \textbf{LSTM-Attention}: The same LSTM architecture augmented with Bahdanau attention~\cite{bahdanau2015neural}, where the output is the attention-weighted sum of the LSTM hidden states.
    \item \textbf{Transformer}: A Transformer encoder following~\cite{iTransformer}, adapted for the Incheon Airport trajectory dataset and equipped with a causal mask to enable autoregressive modeling, where the final memory state serves as the sequence representation.
    \item \textbf{Inverted Transformer}: An Inverted Transformer encoder following~\cite{iTransformer}, adapted for the Incheon Airport trajectory dataset, where the memory states are concatenated and projected into the sequence representation.
    \item \textbf{TCN}: Following the architecture and max-pooled instance representation used in \cite{TS2Vec, InfoTS}, with parameters randomly initialized and no pretraining.
\end{itemize}
All encoders include a linear layer mapping the encoder output to $\mathbf{Z}^{f}_{i,t} \in \mathbb{R}^{320}$, concatenated with the extracted tabular features to yield an 832-dimensional input. This design restricts the model to a single or fixed number of trajectories. The combined features are then fed into a regression bottleneck implemented as a two-layer MLP with 1024 units per layer, ReLU activations, and a dropout rate of 0.1, followed by a final scalar output layer. All neural network-based baselines were trained for 15 epochs using the AdamW optimizer with a learning rate of $1\times10^{-5}$ and the SmoothL1 loss (Eq.~\ref{eq:loss}).

According to Table~\ref{tab:baselinecompare}, a clear performance improvement is observed when moving from tabular-based to trajectory-based models, underscoring the importance of \(X^f_{i,t}\) for $\Delta t_{\text{post},i}$ prediction. Among trajectory-based models, LSTM-Attention performs comparably to the vanilla LSTM. Both Transformer-based models demonstrate stronger performance by preserving long-range dependencies and attending to relevant parts of \(X^f_{i,t}\). Although TCN employs the same architecture as the best-performing TS2Vec encoder, it is trained in a fully supervised manner and performs competitively by directly mapping trajectory semantics to \(\Delta t_{\text{post},i}\). However, trajectory-based models remain inferior to LLM4Delay, indicating that representations learned from \(X^f_{i,t}\) alone, even when optimized with a regression objective, are insufficient without additional operational and airspace context.

\begin{figure*}[h]
    \centering
    \hspace{-10pt}
    \subfloat{
        \includegraphics[width=0.17\textwidth]{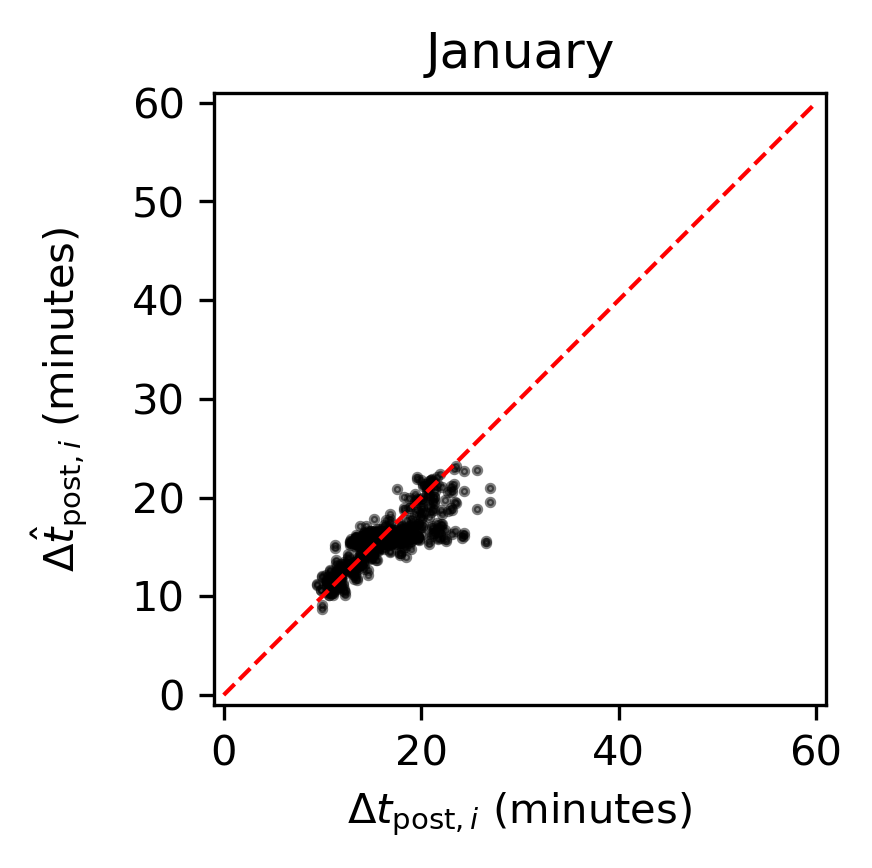}
        \label{fig:rp01}
    }
    \hspace{-13pt}
    \subfloat{
        \includegraphics[width=0.17\textwidth]{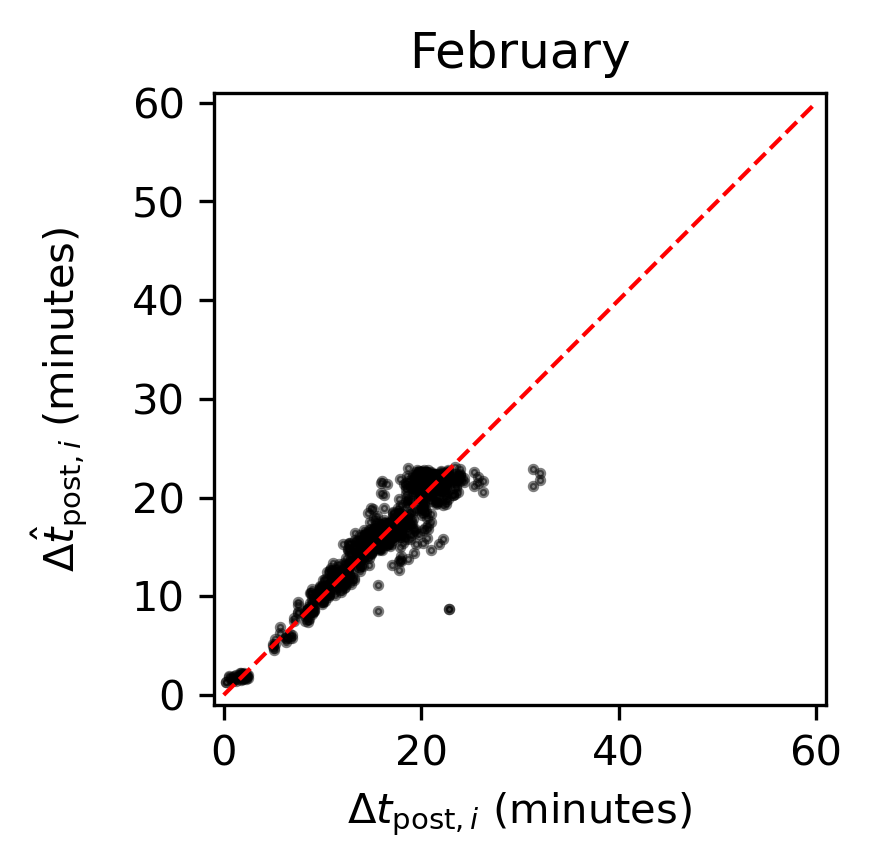}
        \label{fig:rp02}
    }
    \hspace{-13pt}
    \subfloat{
        \includegraphics[width=0.17\textwidth]{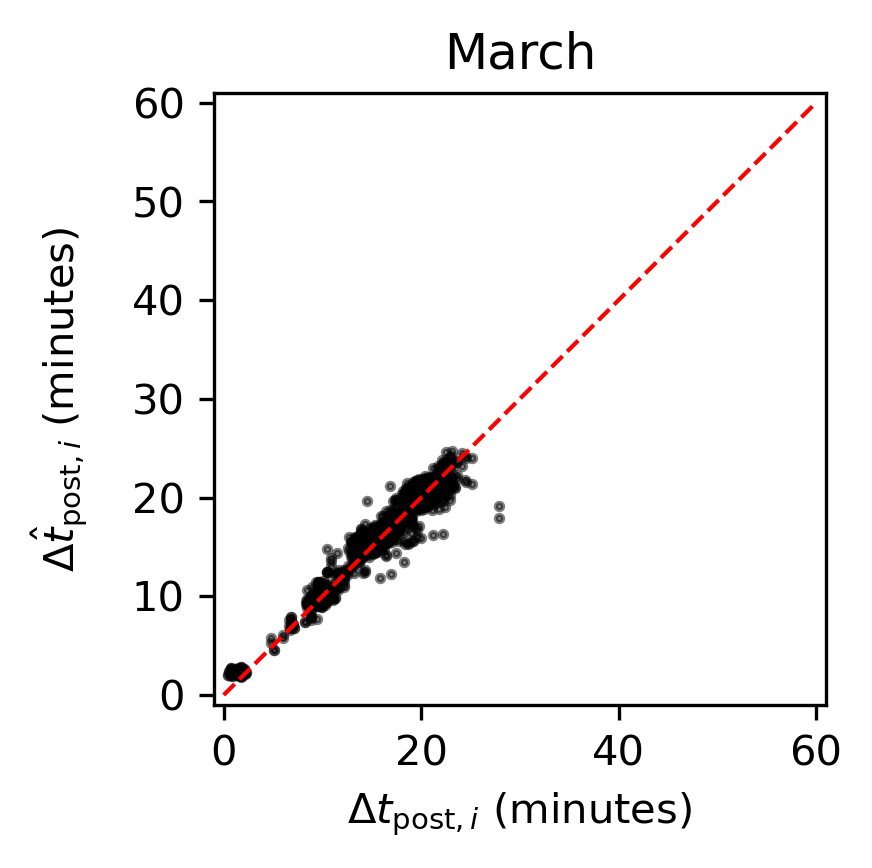}
        \label{fig:rp03}
    }
    \hspace{-13pt}
    \subfloat{
        \includegraphics[width=0.17\textwidth]{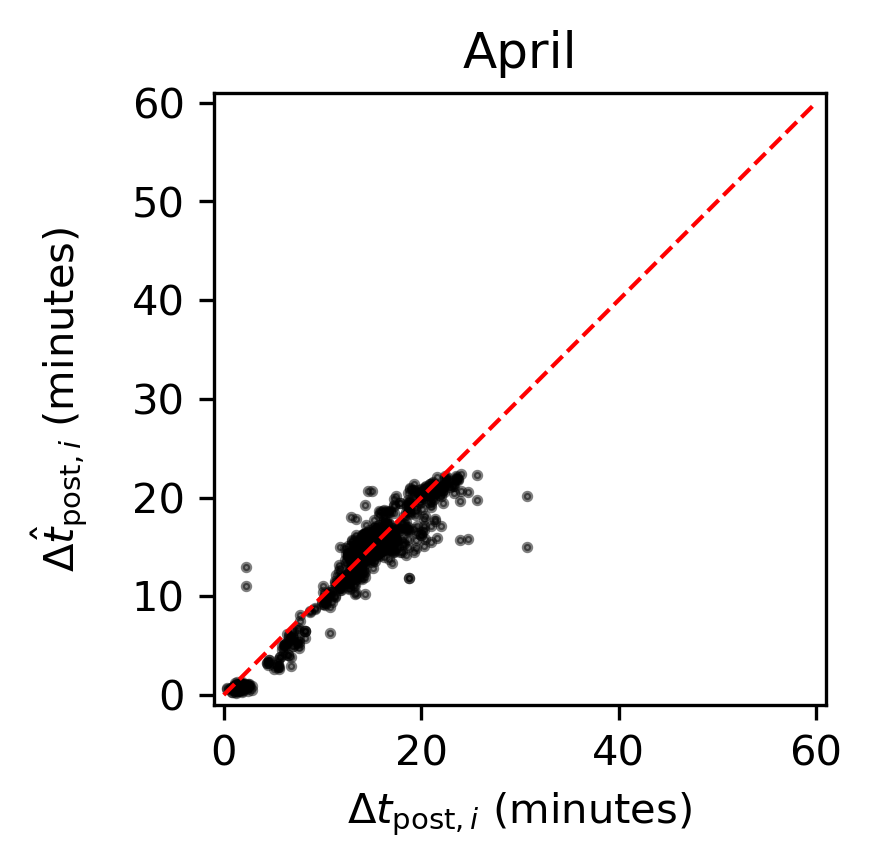}
        \label{fig:rp04}
    }
    \hspace{-13pt}
    \subfloat{
        \includegraphics[width=0.17\textwidth]{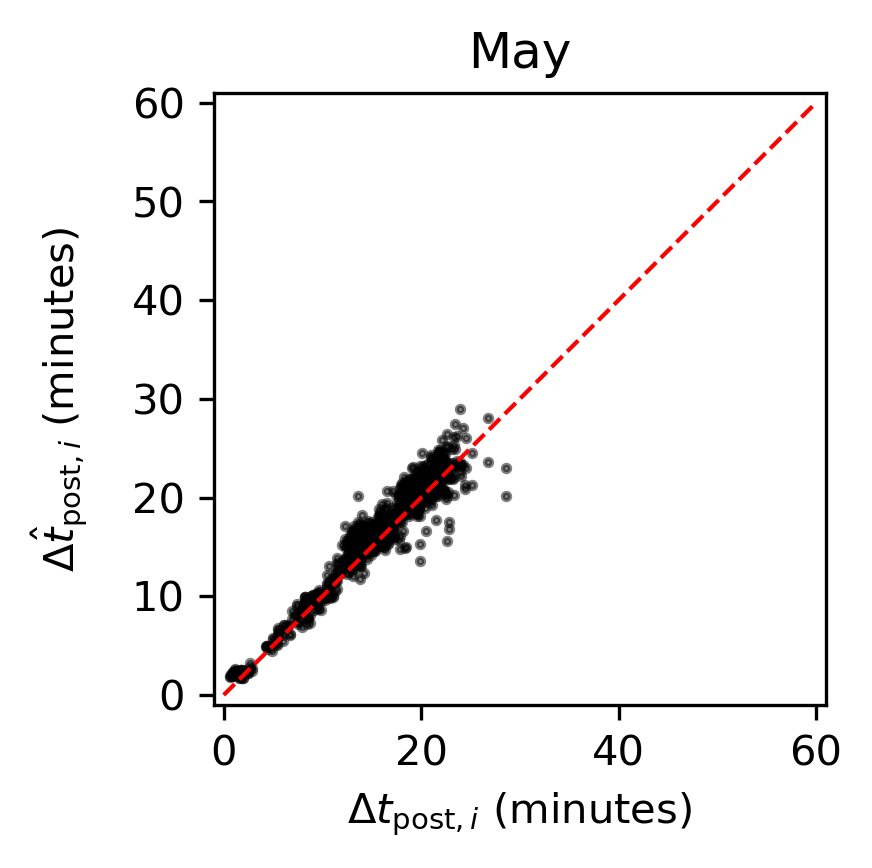}
        \label{fig:rp05}
    }
    \hspace{-13pt}
    \subfloat{
        \includegraphics[width=0.17\textwidth]{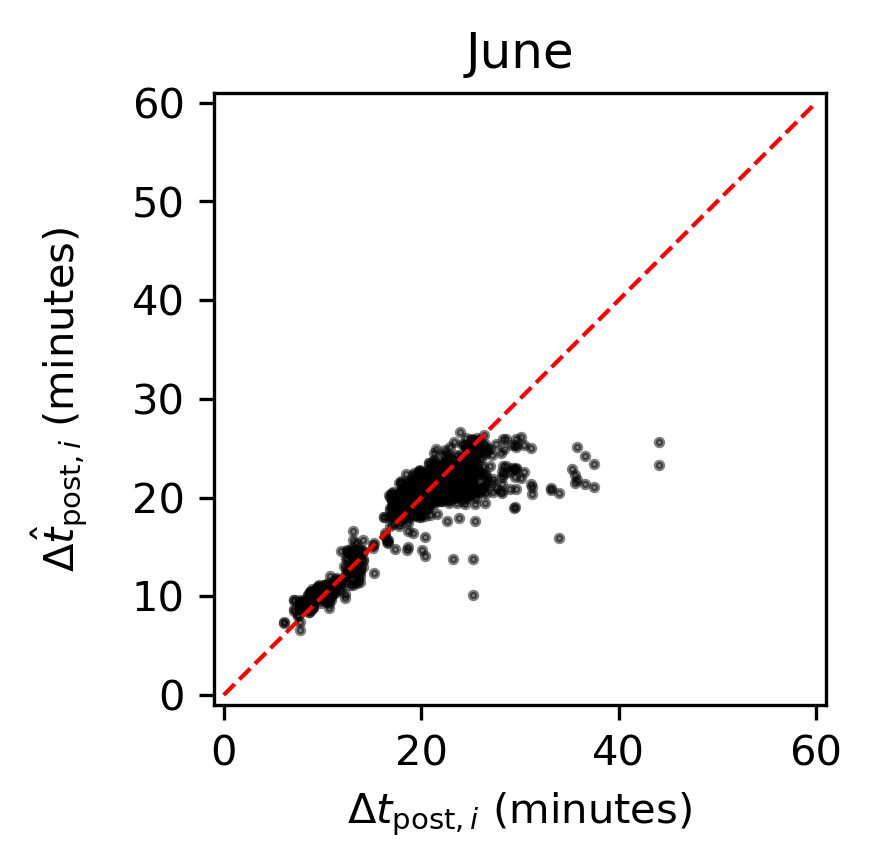}
        \label{fig:rp06}
    }
    
    \vspace{-10pt}
    \hspace{-10pt}
    \subfloat{
        \includegraphics[width=0.17\textwidth]{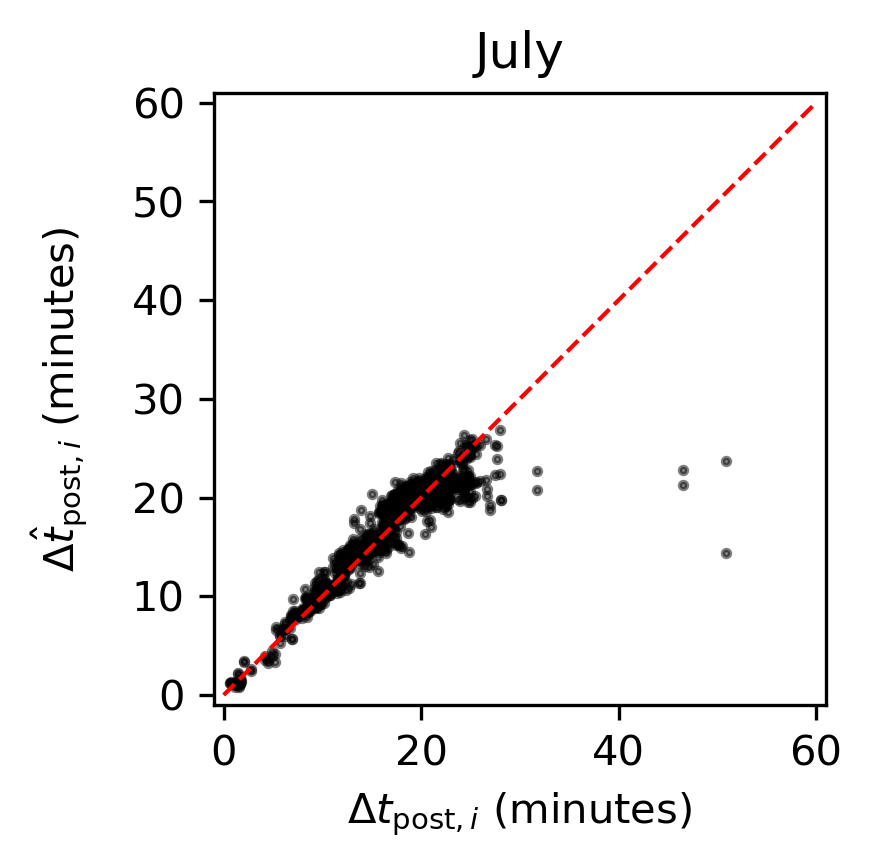}
        \label{fig:rp07}
    }
    \hspace{-13pt}
    \subfloat{
        \includegraphics[width=0.17\textwidth]{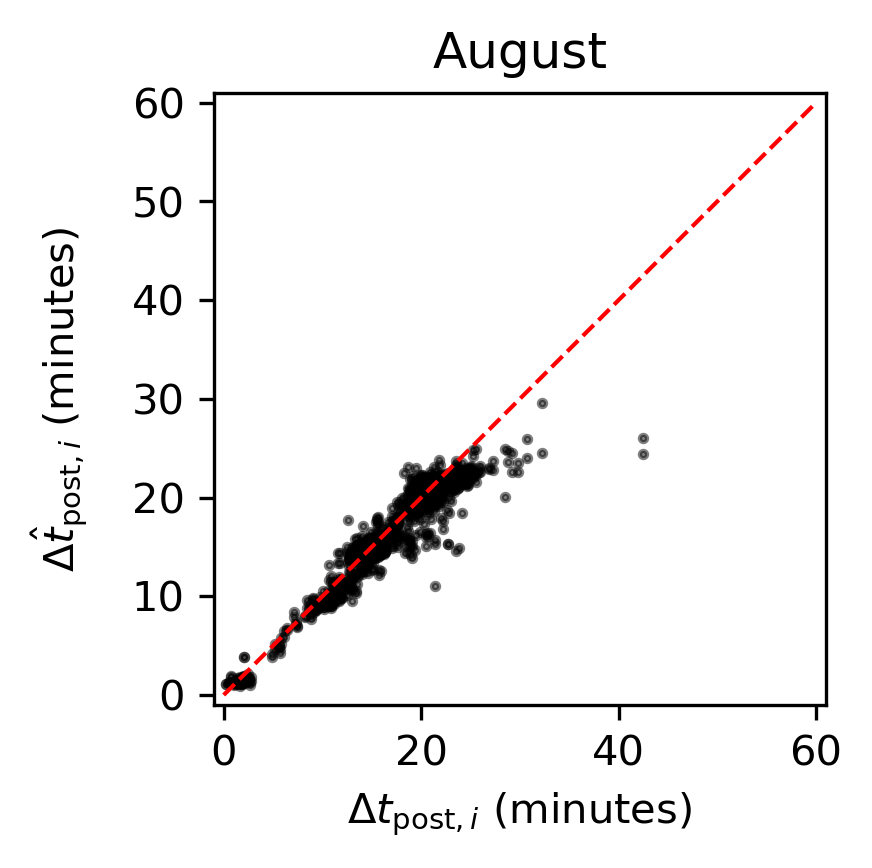}
        \label{fig:rp08}
    }
    \hspace{-13pt}
    \subfloat{
        \includegraphics[width=0.17\textwidth]{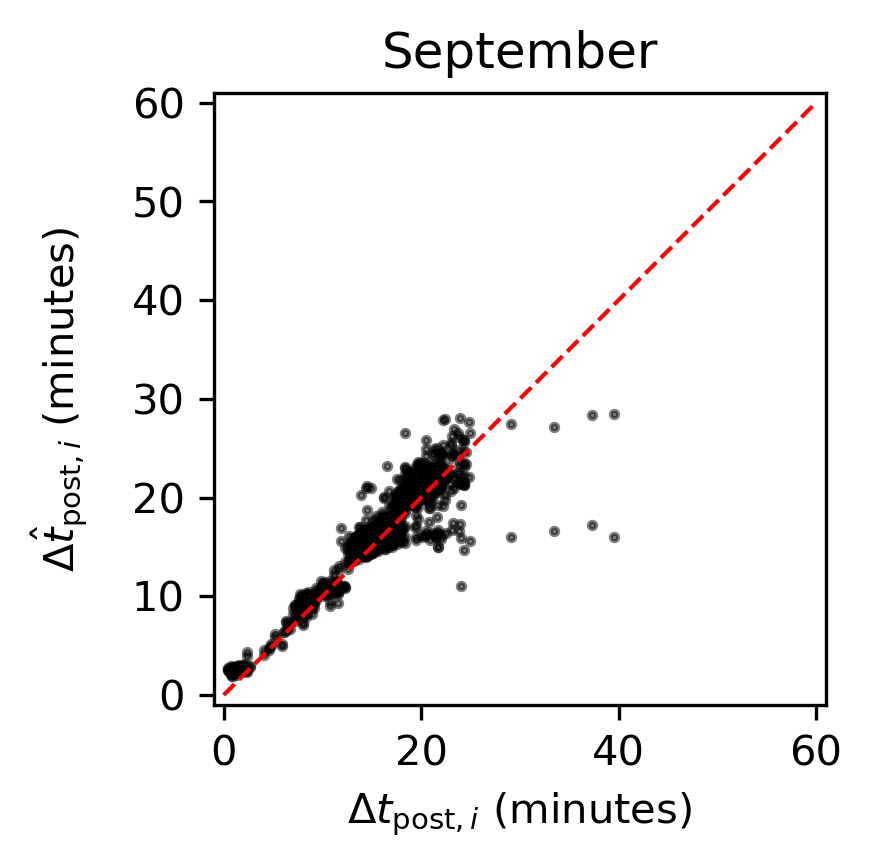}
        \label{fig:rp09}
    }
    \hspace{-13pt}
    \subfloat{
        \includegraphics[width=0.17\textwidth]{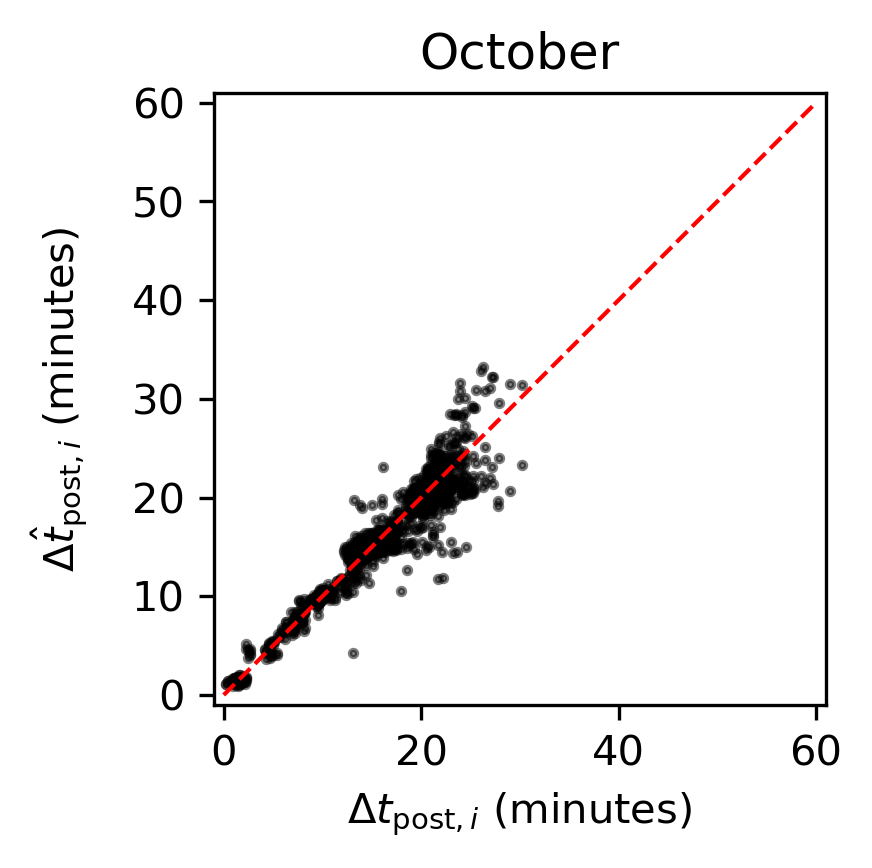}
        \label{fig:rp10}
    }
    \hspace{-13pt}
    \subfloat{
        \includegraphics[width=0.17\textwidth]{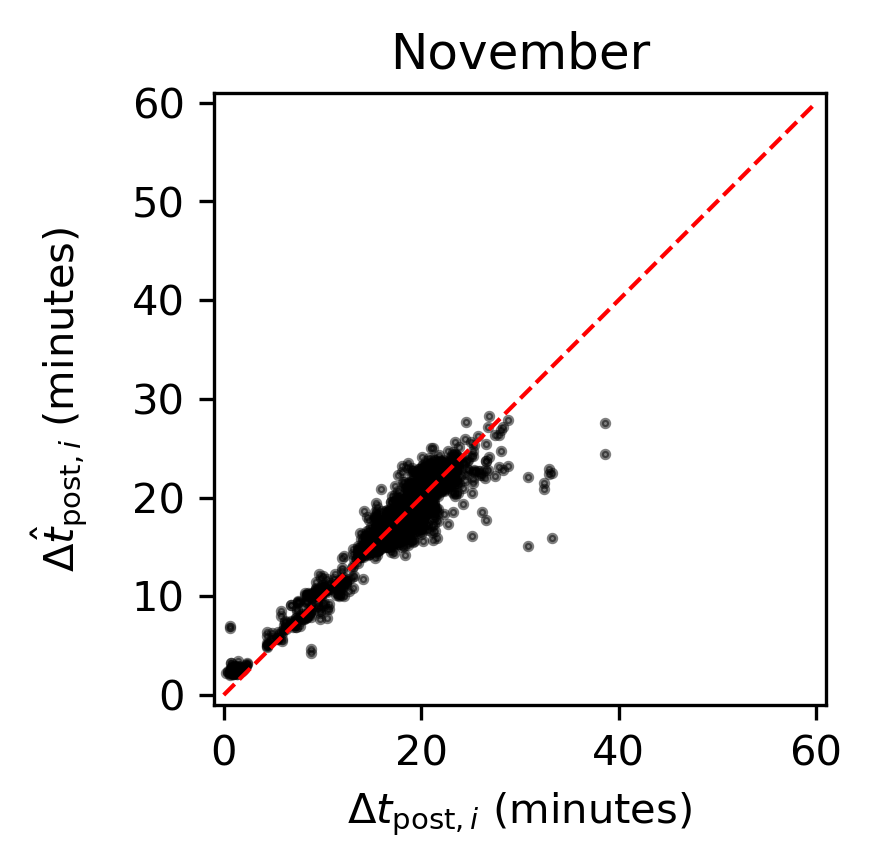}
        \label{fig:rp11}
    }
    \hspace{-13pt}
    \subfloat{
        \includegraphics[width=0.17\textwidth]{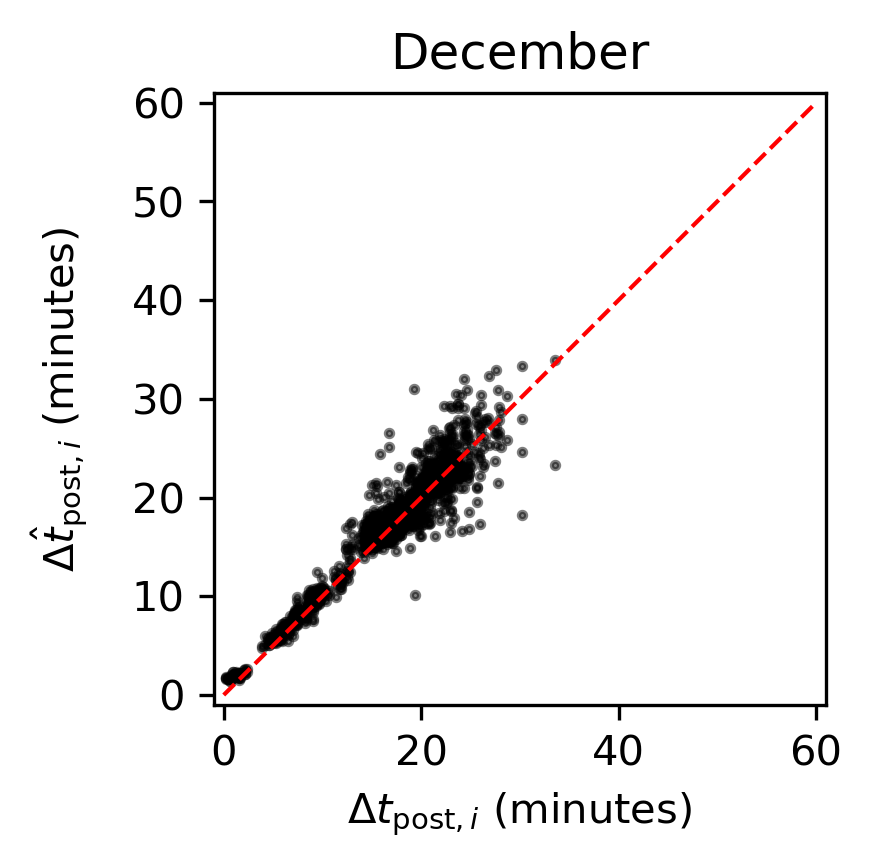}
        \label{fig:rp12}
    }
    \caption{Monthly parity plots of predicted versus ground-truth $\Delta t_{\text{post},i}$ for LLM4Delay on the test dataset}
    \label{fig:parity}
\end{figure*}

LLM4Delay encodes a comprehensive set of multimodal inputs in a unified token-based format, incorporating TAF in $P^{\text{T}}_t$, NOTAMs in $P^{\text{N}}_t$, and additional trajectory information via $X^a_{i,t}$ and $X^p_{i,t}$. These sources provide weather information, operational constraints, and airspace awareness that influence $\Delta t_{\text{post},i}$. As a result, LLM4Delay consistently achieves the best performance across all monthly datasets when equipped with the TS2Vec encoder and Pythia-1B. Notably, this performance is achieved despite the framework not being fully trainable, since it leverages the generalizability of both the pretrained trajectory encoder and the pretrained LLM. The parity plot in Fig.~\ref{fig:parity} further shows strong agreement between predicted and actual values on the test data, with predictions closely aligning with the ideal line.

\clearpage
\begin{table*}
\centering
\rotatebox{90}{%
\begin{minipage}{1.2\textwidth}
\caption{Performance of LLM4Delay Compared to Existing Baseline Frameworks in Air Traffic Management}
\resizebox{\textwidth}{!}{
\begin{tabular}{llrrrrrrrrrrrrr}
\toprule
& & \multicolumn{13}{c}{\textbf{Datasets}} \\
\cmidrule(lr){3-15}
\textbf{Metrices} & \textbf{Baseline Frameworks} & \textbf{JAN} & \textbf{FEB} & \textbf{MAR} & \textbf{APR} & \textbf{MAY} & \textbf{JUN} & \textbf{JUL} & \textbf{AUG} & \textbf{SEP} & \textbf{OCT} & \textbf{NOV} & \textbf{DEC} & \textbf{AVG}\\
\midrule
\multirow{14}{*}{\textbf{MAE\textsuperscript{\(\downarrow\)}}}
&$F^{\text{F}}_i, F^{\text{M}}_t, N^a_{i,t}$ &&&&&&&&&&&&& \\
&Linear Regression & 2.3288 & 5.0626 & 3.6065 & 4.6326 & 2.9715 & 4.1762 & 2.7224 & 3.5705 & 3.7327 & 3.3628 & 3.6959 & 3.3840 & 3.6039 \\
&SVM               & 3.3362 & 3.9226 & 4.4374 & 3.6143 & 2.7354 & 4.3150 & 3.1979 & 3.7167 & 4.5773 & 3.6128 & 3.6603 & 3.1404 & 3.6889 \\
&Random Forest     & 1.8698 & 3.9847 & 2.6981 & 2.7696 & 2.5018 & 3.9416 & 3.1782 & 4.1258 & 2.5397 & 2.7006 & 2.2421 & 2.5549 & 2.9256 \\
&XGBoost           & 1.5301 & 4.0814 & 2.7675 & 2.6433 & 2.8061 & 3.6868 & 2.6081 & 3.3401 & 2.7166 & 2.6616 & 2.3605 & 2.4543 & 2.8047 \\
&MLP               & 2.1088 & 3.9551 & 4.0178 & 3.3816 & 2.4848 & 3.3094 & 2.5511 & 3.4941 & 3.9842 & 3.2098 & 3.1997 & 2.7108 & 3.2006 \\
\cmidrule(lr){2-15}
&$F^{\text{F}}_i, F^{\text{M}}_t, N^a_{i,t}, X^{f}_{i,t}$ &&&&&&&&&&&&&\\
&LSTM                 & 1.8450 & 1.8376 & 1.5706 & 1.4682 & 1.2970 & 1.9256 & 1.3615 & 1.4797 & 1.8761 & 1.3822 & 1.4519 & 1.2411 & 1.5614 \\
&LSTM-Attention       & 1.5927 & 1.9572 & 1.6242 & 1.5456 & 1.3693 & 2.0494 & 1.5128 & 1.6129 & 2.1349 & 1.4748 & 1.6919 & 1.2573 & 1.6519 \\
&Transformer          & \underline{1.5179} & 1.7082 & 1.4219 & 1.5921 & 1.2185 & 1.7882 & 1.3426 & 1.4932 & 1.6156 & \underline{1.2592} & 1.4193 & 1.2135 & 1.4659 \\
&Inverted Transformer & 1.5619 & \underline{1.6226} & 1.4878 & 1.4415 & 1.3059 & 1.7794 & 1.3094 & 1.4946 & 1.4503 & 1.2988 & 1.4897 & 1.2412 & 1.4569 \\
&TCN                  & 2.2473 & 1.7293 & \underline{1.2566} & \underline{1.2388} & \underline{1.0689} & \underline{1.6767} & \textbf{1.1582} & \underline{1.3696} & \underline{1.4118} & 1.2786 & \underline{1.1896} & \underline{1.1215} & \underline{1.3956} \\
\cmidrule(lr){2-15}
&$P^{\text{F}}_{i,t}, P^{\text{M}}_{t}, P^{\text{T}}_{t}, P^{\text{N}}_{t}, X^f_{i,t}, X^a_{i,t}, X^p_{i,t}$ &&&&&&&&&&&&& \\
&LLM4Delay (Ours) & \textbf{1.3134} & \textbf{1.1296} & \textbf{0.9468} & \textbf{1.0248} & \textbf{1.0370} & \textbf{1.4994} & \underline{1.1790} & \textbf{1.0162} & \textbf{1.1470} & \textbf{1.0341} & \textbf{1.1344} & \textbf{1.0517} & \textbf{1.1261} \\
\midrule
\multirow{14}{*}{\textbf{MSE\textsuperscript{\(\downarrow\)}}}
&$F^{\text{F}}_i, F^{\text{M}}_t, N^a_{i,t}$ &&&&&&&&&&&&& \\
&Linear Regression & 10.5743 & 37.8387 & 21.6914 & 36.2099 & 15.3911 & 27.6779 & 15.1147 & 21.8087 & 23.4525 & 18.9601 & 21.4376 & 19.1653 & 22.4435 \\
&SVM               & 16.1035 & 24.6755 & 30.5925 & 22.3557 & 13.7995 & 31.1163 & 18.1664 & 24.5053 & 33.7065 & 19.4848 & 23.1674 & 17.0171 & 22.8909 \\
&Random Forest     & 7.0316  & 27.0162 & 12.5990 & 16.5707 & 11.2390 & 26.5872 & 18.5718 & 26.9343 & 13.8467 & 13.3552 & 10.0289 & 12.2379 & 16.3349 \\
&XGBoost           & 4.7279  & 29.2474 & 12.8970 & 14.2009 & 12.4419 & 24.5775 & 13.4513 & 18.2157 & 13.6410 & 11.8070 & 10.4083 & 10.8973 & 14.7094 \\
&MLP               & 7.8610  & 24.7174 & 25.8329 & 21.1614 & 11.6389 & 20.3649 & 13.2979 & 22.0848 & 27.0943 & 17.5511 & 17.9052 & 12.9063 & 18.5347 \\
\cmidrule(lr){2-15}
&$F^{\text{F}}_i, F^{\text{M}}_t, N^a_{i,t}, X^{f}_{i,t}$ &&&&&&&&&&&&& \\
&LSTM                 & 5.9152 & 7.1766 & 4.2926 & 3.9092 & 3.0825 & 9.0215  & 5.6562 & 4.7723 & 6.2364 & 3.9711 & 4.6929 & 3.4248 & 5.1793 \\
&LSTM-Attention       & 4.9053 & 7.7931 & 4.5120 & 4.1649 & 3.3592 & 10.4593 & 6.2982 & 4.9553 & 7.9232 & 4.4417 & 5.7616 & 3.6865 & 5.6884 \\
&Transformer          & \underline{4.4415} & 5.7577 & 3.3427 & 4.8667 & 2.6303 & 7.7314  & 4.9695 & 4.6650 & 5.1773 & \underline{3.2886} & 4.3975 & 3.2935 & 4.5468 \\
&Inverted Transformer & 4.6605 & \underline{5.2572} & 3.6432 & 3.9300 & 3.0164 & 7.7290  & 5.5960 & 4.7767 & 4.6546 & 3.5857 & 5.0118 & 3.5114 & 4.6144 \\
&TCN                  & 7.8404 & 6.3228 & \underline{2.9137} & \underline{3.1690} & \underline{2.1717} & \underline{7.6281}  & \underline{4.6671} & \underline{3.9255} & \underline{4.2207} & 3.6307 & \underline{3.3362} & \underline{2.9935} & \underline{4.4016} \\
\cmidrule(lr){2-15}
&$P^{\text{F}}_{i,t}, P^{\text{M}}_{t}, P^{\text{T}}_{t}, P^{\text{N}}_{t}, X^f_{i,t}, X^a_{i,t}, X^p_{i,t}$ &&&&&&&&&&&&& \\
&LLM4Delay (Ours) & \textbf{3.1676} & \textbf{3.0012} & \textbf{1.6113} & \textbf{2.3870} & \textbf{1.9398} & \textbf{6.5747} & \textbf{4.6274} & \textbf{2.2915} & \textbf{3.7346} & \textbf{2.6495} & \textbf{2.9348} & \textbf{2.5216} & \textbf{3.1201} \\
\midrule
\multirow{14}{*}{\textbf{R\(^2_{\Delta t}\)\textsuperscript{\(\uparrow\)}}}
&$F^{\text{F}}_i, F^{\text{M}}_t, N^a_{i,t}$ &&&&&&&&&&&&& \\
&Linear Regression & -0.3821 & -0.4310 &  0.2684 & -0.3912 & 0.4175 &  0.0338 & 0.4370 & 0.2733 &  0.2355 & 0.3662 & 0.2359 & 0.2495 & 0.1094 \\
&SVM               & -1.1048 &  0.0668 & -0.0318 &  0.1411 & 0.4778 & -0.0862 & 0.3233 & 0.1835 & -0.0987 & 0.3486 & 0.1743 & 0.3336 & 0.0606 \\
&Random Forest     &  0.0809 & -0.0217 &  0.5751 &  0.3633 & 0.5747 &  0.0719 & 0.3082 & 0.1026 &  0.5486 & 0.5535 & 0.6426 & 0.5207 & 0.3600 \\
&XGBoost           &  0.3820 & -0.1061 &  0.5650 &  0.4544 & 0.5291 &  0.1420 & 0.4989 & 0.3931 &  0.5554 & 0.6053 & 0.6290 & 0.5732 & 0.4351 \\
&MLP               & -0.0275 &  0.0652 &  0.1288 &  0.1869 & 0.5595 &  0.2891 & 0.5047 & 0.2641 &  0.1168 & 0.4133 & 0.3618 & 0.4946 & 0.2798 \\
\cmidrule(lr){2-15}
&$F^{\text{F}}_i, F^{\text{M}}_t, N^a_{i,t}, X^{f}_{i,t}$ &&&&&&&&&&&&& \\
&LSTM                 & 0.2269  & 0.7286 & 0.8552 & 0.8498 & 0.8833 & 0.6851 & 0.7893 & 0.8410 & 0.7967 & 0.8672 & 0.8327 & 0.8659 & 0.7685 \\
&LSTM-Attention       & 0.3589  & 0.7053 & 0.8478 & 0.8400 & 0.8729 & 0.6349 & 0.7654 & 0.8349 & 0.7417 & 0.8515 & 0.7946 & 0.8556 & 0.7586 \\
&Transformer          & \underline{0.4195}  & 0.7822 & 0.8873 & 0.8130 & 0.9005 & 0.7301 & 0.8149 & 0.8446 & 0.8312 & \underline{0.8901} & 0.8433 & 0.8710 & \underline{0.8023} \\
&Inverted Transformer & 0.3908  & \underline{0.8012} & 0.8771 & 0.8490 & 0.8858 & 0.7302 & 0.7915 & 0.8408 & 0.8483 & 0.8801 & 0.8214 & 0.8625 & 0.7982 \\
&TCN                  & -0.0248 & 0.7609 & \underline{0.9017} & \underline{0.8782} & \underline{0.9178} & \underline{0.7337} & \underline{0.8262} & \underline{0.8692} & \underline{0.8624} & 0.8786 & \underline{0.8811} & \underline{0.8828} & 0.7807 \\
\cmidrule(lr){2-15}
&$P^{\text{F}}_{i,t}, P^{\text{M}}_{t}, P^{\text{T}}_{t}, P^{\text{N}}_{t}, X^f_{i,t}, X^a_{i,t}, X^p_{i,t}$ &&&&&&&&&&&&& \\
&LLM4Delay (Ours) & \textbf{0.5860} & \textbf{0.8865} & \textbf{0.9457} & \textbf{0.9083} & \textbf{0.9266} & \textbf{0.7705} & \textbf{0.8276} & \textbf{0.9236} & \textbf{0.8783} & \textbf{0.9114} & \textbf{0.8954} & \textbf{0.9012} & \textbf{0.8634} \\
\bottomrule
\multicolumn{15}{l}{\footnotesize \textsuperscript{\(\downarrow\)} Lower is better, \textsuperscript{\(\uparrow\)} Higher is better; LLM4Delay is equipped with TS2Vec and Pythia-1B.}
\label{tab:baselinecompare}
\end{tabular}
}
\end{minipage}
}
\end{table*}
\clearpage

\subsection{Comparison Study with Existing Adaptation Techniques}
One of our core contributions is an effective cross-modality adaptation method that maps trajectory time-series data to the language modality for accurate delay prediction. To assess its effectiveness, this section compares our approach with existing time-series-to-language adaptation techniques. Following our evaluation protocol, we replaced the adaptation module, consisting of \(f_{\text{enc}}(\cdot)\) and \(f_{\text{xa}}(\cdot)\), with the techniques visualized in Fig.~\ref{fig:fusion}. We first consider non-learnable approaches:
\begin{itemize}
    \item \textbf{Tokenization}: Following LLMTIME~\cite{llmtime}, $X^f_{i,t}$, $X^a_{i,t}$, and $X^p_{i,t}$ are serialized via $S(\cdot)$, tokenized, and flattened along the sequence dimension, allowing the LLM to attend to all trajectories. Combined context length is capped at 2048 tokens. The trajectory-informed embedding is given by:
\end{itemize}
\begin{align}
    \mathbf{Z}^{T}_{i,t} = \text{Concat}\big[&\text{Emb}[\text{T}(P_{st,1})], \text{Emb}[\text{T}(S(X^f_{i,t}))],\ \text{Emb}[\text{T}(P_{st,2})],\nonumber\\
    &\ \text{Emb}[\text{T}(S(X^a_{i,t}))],\ \text{Emb}[\text{T}(P_{st,3})], \text{Emb}[\text{T}(S(X^p_{i,t}))],\ \text{Emb}[\text{T}(P_{st,4})]\big].
    \label{eq:trajinform}
\end{align}

Learnable baselines do not natively support multiple trajectories, so for each $s_{i,t}$, $X^f_{i,t}$, $X^a_{i,t}$, and $X^p_{i,t}$ are concatenated into a single unified series $X_{i,t}$. As they also do not handle unequal-length segments, zero-padding is inserted between trajectories and at the sequence end to separate them and ensure sub-series divisibility:

\begin{equation}
\begin{aligned}
X_{i,t}
= \mathrm{Concat}\big[
&X^f_{i,t}, \mathbf{0}_{\mathrm{pad}}, \\
&X^a_{i,t,1},\mathbf{0}_{\mathrm{pad}},X^a_{i,t,2},\mathbf{0}_{\mathrm{pad}},\dots,X^a_{i,t,N^a_{i,t}},\mathbf{0}_{\mathrm{pad}}, \\
&X^p_{i,t,1},\mathbf{0}_{\mathrm{pad}},X^p_{i,t,2},\mathbf{0}_{\mathrm{pad}},\dots,X^p_{i,t,N^p_{i,t}},\mathbf{0}_{\mathrm{pad}}
\big],
\end{aligned}
\end{equation}
where \(X_{i,t} \in \mathbb{R}^{\mathbb{T} \times 9} \) and \(\mathbb{T}=T^f_{i,t} + \sum_{j=1}^{N^a_{i,t}} T^a_{i,t/j} + \sum_{j=1}^{N^p_{i,t}} T^p_{i,t/j} + \sum_{j=1}^{N^a_{i,t} + N^p_{i,t} +1} T_{\mathrm{pad},j}\). We consider the following embedding methods for $f_{\text{emb}}(\cdot)$:
\begin{itemize}
    \item \textbf{Patch Reprogramming}: We adopted the patch reprogramming layer from TimeLLM~\cite{timellm}, using the model hyperparameters specified for the LTF-ILI dataset.
    \item \textbf{Segment Embedding}: Following AutoTimes~\cite{autotimes}, segments are embedded using an MLP with default hyperparameters from the public implementation, with positional encoding supplied by the LLM backbone.
\end{itemize} 
For learnable approaches, segment length is set to 96. Under channel independence~\cite{PatchTST}, the 9 features in $X_{i,t}$ (Eq.~\ref{eq:trajfeature}) are treated as independent univariate time series. The embeddings are concatenated into a single sequence, preceded by the prompt $P_{N^a_{i,t},N^p_{i,t}}$, which encodes $N^a_{i,t}$ and $N^p_{i,t}$ as in \cite{timellm}, allowing the LLM backbone to attend jointly to embeddings from all features. The trajectory-informed embedding is:
\begin{equation}
\begin{aligned}
\mathbf{Z}^T_{i,t}
= \mathrm{Concat}\big[
    &\text{Emb}\!\big[\text{T}(P_{N^a_{i,t},N^p_{i,t}})\big], \\
    &f_{\text{emb}}\!\big(X_{i,t}^{(1)}\big),\;
     f_{\text{emb}}\!\big(X_{i,t}^{(2)}\big),\; \dots,\;
     f_{\text{emb}}\!\big(X_{i,t}^{(9)}\big), \\
    &\text{Emb}\!\big[\text{T}(P_{st,4})\big]
\big].
\end{aligned}
\end{equation}

Comparative results are presented in Table~\ref{tab:fusioncompare}. Tokenization relies on non-learnable input serialization, requiring the LLM to infer time-series semantics from low-level digit tokens. Learnable baselines (patch reprogramming, segment embedding) improve on this by adapting feature-wise segment-level semantics rather than digit-level semantics, but still rely on the LLM to infer global trajectory meaning across segments, limiting their applicability to \(\Delta t_{\text{post},i}\) prediction. LLM4Delay instead leverages pre-encoded instance-level semantics from multiple trajectories via instance-level projection, avoiding the need for the LLM to reconstruct temporal structure from fragmented inputs. This yields the best performance, supporting our hypothesis that accurate \(\Delta t_{\text{post},i}\) estimation depends on instance-level trajectory semantics capturing underlying maneuvering procedures.

However, instance-level projection relies on a pretrained $f_{\text{enc}}(\cdot)$, whose pretraining data may not be available for all airports; although available for Incheon Airport via ATFMTraj~\cite{atfmtraj}, new airports would require separate data curation. Even with data available, effective representation-learning techniques remain an important component of training $f_{\text{enc}}(\cdot)$, as discussed in the following section.

\subsection{Ablation Study of Trajectory Encoders}
This section evaluates LLM4Delay with the frozen \(f_{\text{enc}}(\cdot)\) replaced by candidate representation learning frameworks pretrained on the ATFMTraj~\cite{atfmtraj} Incheon Airport dataset, highlighting the encoder's impact on cross-modality adaptation and supporting our choice of representation learning method. Reproduction details for each \(f_{\text{enc}}(\cdot)\) are as follows:
\begin{itemize}
    \item \textbf{TCN-AE}: We adopt a TCN architecture, as in \cite{TS2Vec}, for both encoder and decoder, following \cite{TCNAE_Thill}, with max-pooling and repetition-based upsampling, trained using the MSE loss.
    \item \textbf{TS2Vec}~\cite{TS2Vec}: learns time-series representations via hierarchical contrastive learning; we followed the authors' reproduction settings for the UCR/UEA datasets.
    \item \textbf{InfoTS}~\cite{InfoTS}: applies meta-learning to select augmentations; we reproduced the same hyperparameters reported for the UCR/UEA datasets.
    \item \textbf{ATSCC}~\cite{ATSCC}: employs segmentation-based contrastive learning; strictly following the paper, we grid-searched the segmentation threshold and loss temperature, fixing other hyperparameters.
\end{itemize}

\begin{figure}[h]
    \centering
    \vspace{-15pt}

    \subfloat[TCN-AE]{
        \includegraphics[width=0.22\linewidth]{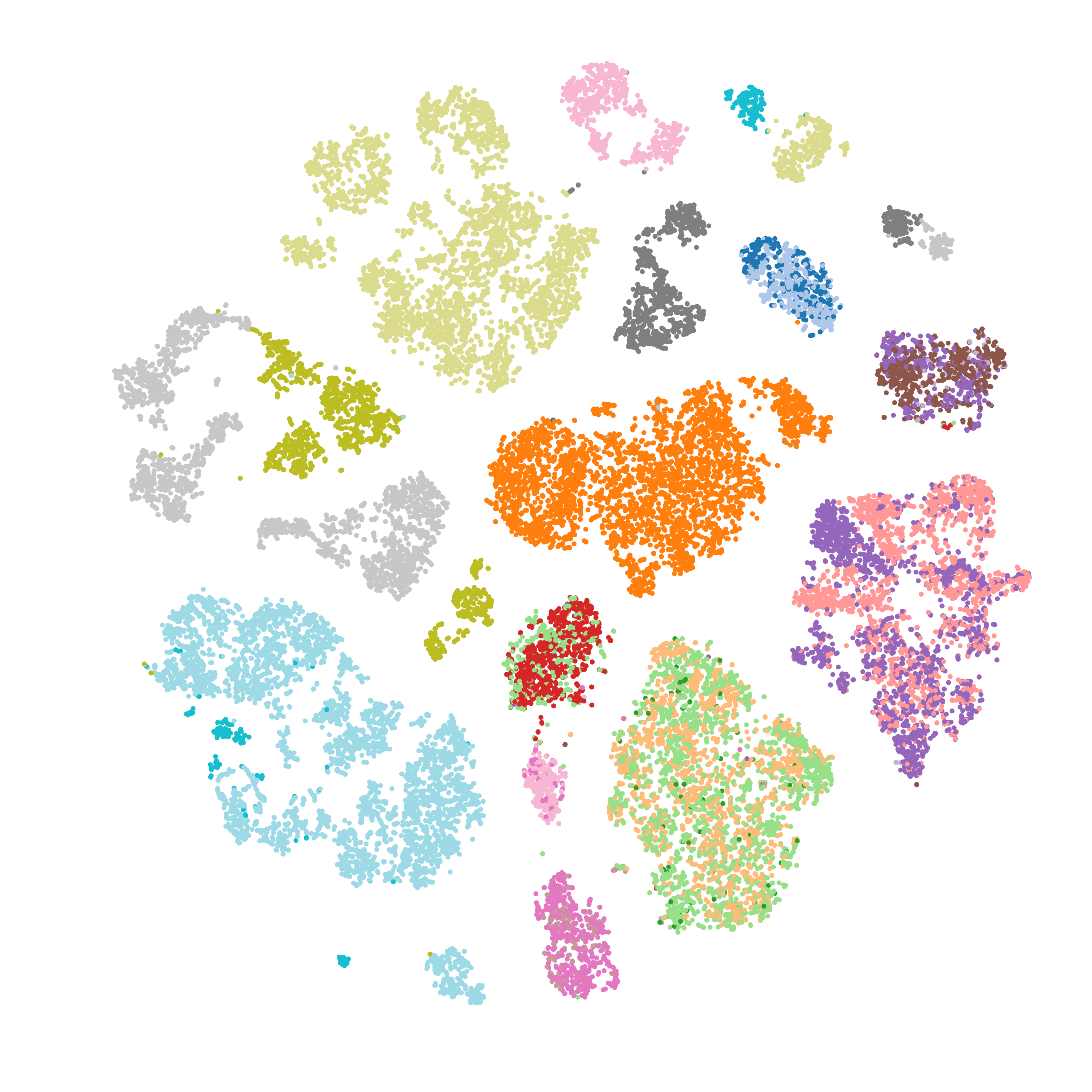}
        \label{fig:TCNAE}
    }
    \subfloat[TS2Vec]{
        \includegraphics[width=0.22\linewidth]{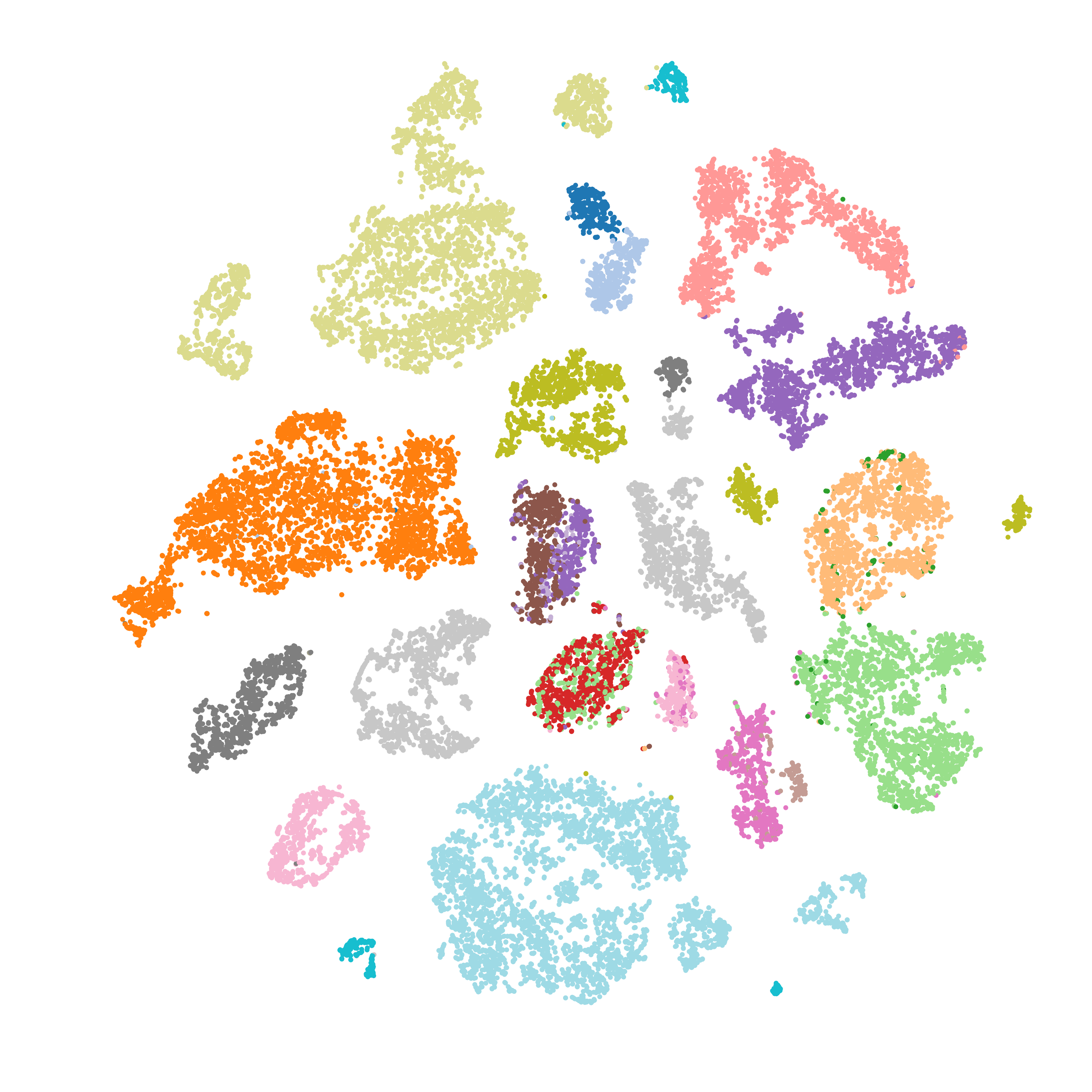}
        \label{fig:TS2Vec}
    }
    \subfloat[InfoTS]{
        \includegraphics[width=0.22\linewidth]{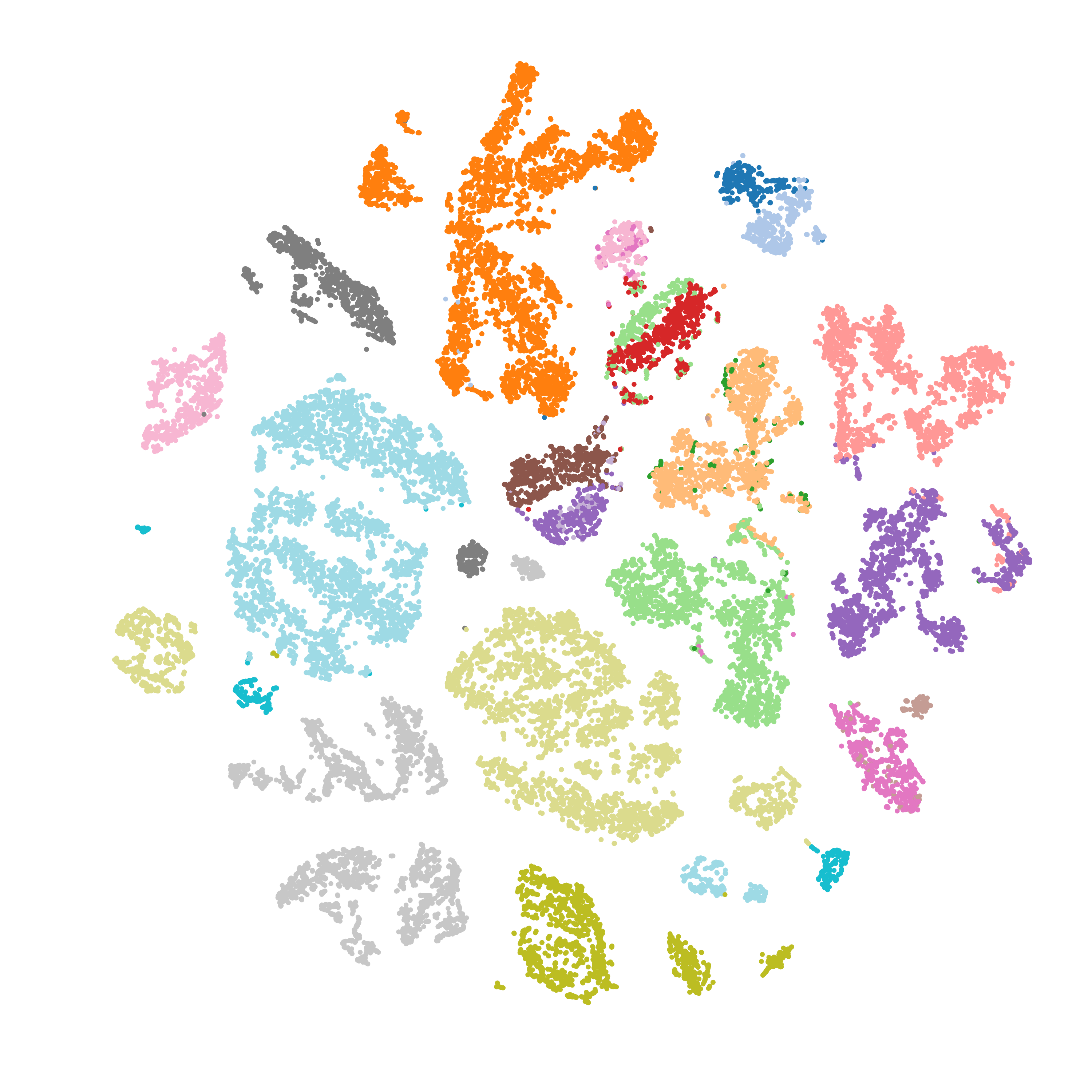}
        \label{fig:InfoTS}
    }
    \subfloat[ATSCC]{
        \includegraphics[width=0.22\linewidth]{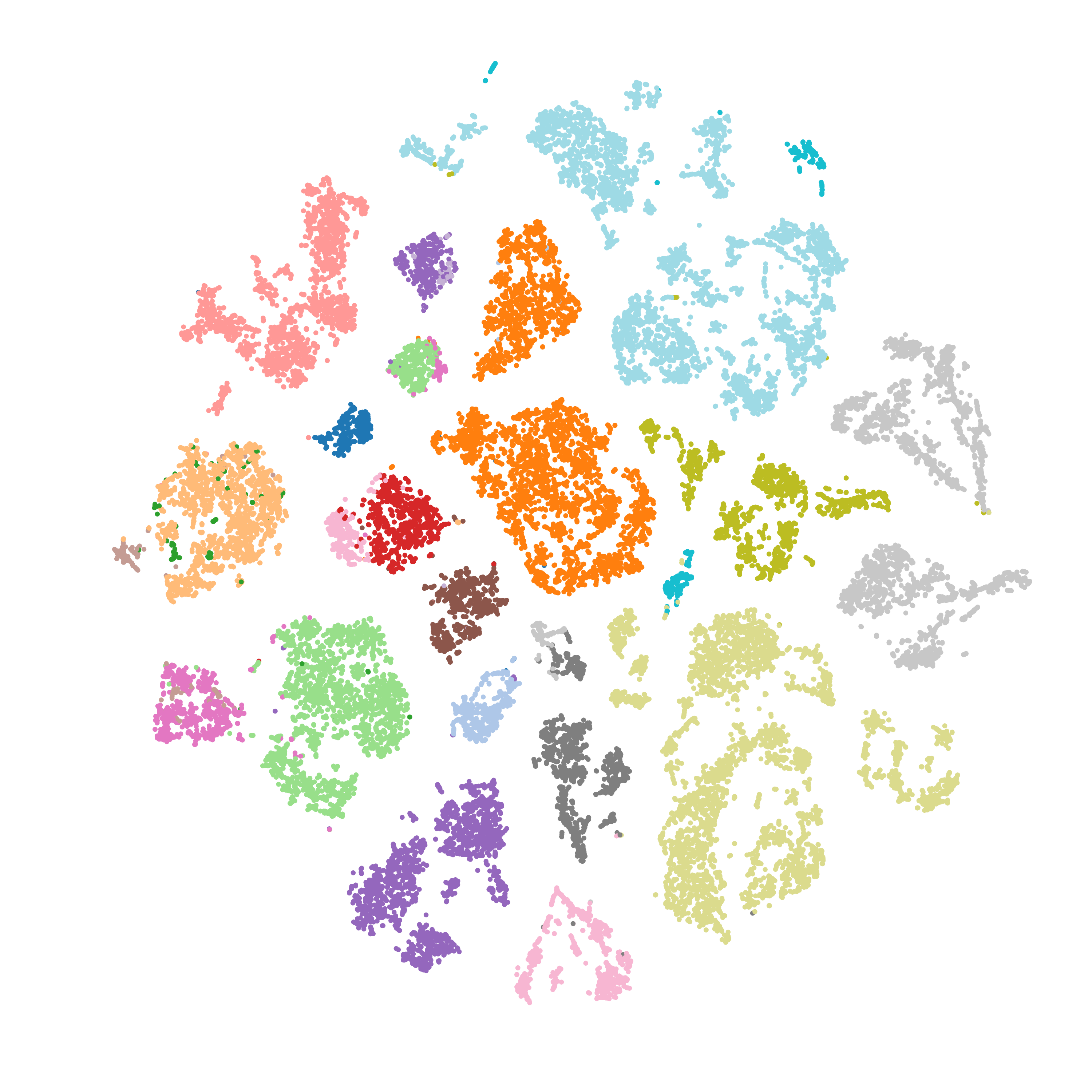}
        \label{fig:ATSCC}
    }

    \caption{t-SNE visualization of learned representations on the ATFMTraj Incheon Airport pretraining dataset.}
    \label{fig:tsne}
\end{figure}

Fig.~\ref{fig:tsne} shows t-SNE plots of instance-level trajectory embeddings from the prepared encoders on the pretraining dataset. Class separation indicates the encoders capture instance-level semantics aligned with maneuvering procedures, with some runway-related ambiguity remaining.

\clearpage
\begin{table*}
\centering
\rotatebox{90}{%
\begin{minipage}{1.2\textwidth}
\caption{Performance of LLM4Delay Compared to Existing Cross-Modality Adaptation Techniques}
\resizebox{\textwidth}{!}{
\begin{tabular}{llrrrrrrrrrrrrr}
\toprule
& & \multicolumn{13}{c}{\textbf{Datasets}} \\
\cmidrule(lr){3-15}
\textbf{Metrices} & \textbf{Baseline Adaptations} & \textbf{JAN} & \textbf{FEB} & \textbf{MAR} & \textbf{APR} & \textbf{MAY} & \textbf{JUN} & \textbf{JUL} & \textbf{AUG} & \textbf{SEP} & \textbf{OCT} & \textbf{NOV} & \textbf{DEC} & \textbf{AVG}\\
\midrule
\multirow{4}{*}{\textbf{MAE\textsuperscript{\(\downarrow\)}}}
&Tokenization        & 2.4569 & 3.3775 & 3.6813 & 3.1706 & 3.0406 & 3.2649 & 3.4521 & 3.0392 & 3.3507 & 3.3281 & 2.7655 & 3.6929 & 3.2184 \\
&Patch Reprogramming & 2.1771 & 3.6663 & 2.7218 & 2.2217 & 2.3924 & 3.0006 & 2.7047 & 2.2651 & 2.9557 & 2.4446 & 3.0968 & 1.7600 & 2.6172 \\
&Segment Embedding   & \underline{2.0592} & \underline{1.9910} & \underline{1.3897} & \underline{1.4285} & \underline{1.3471} & \underline{1.9210} & \underline{1.5156} & \underline{1.3022} & \underline{1.3922} & \underline{1.2521} & \underline{1.3985} & \underline{1.3005} & \underline{1.5248} \\
&Instance-level Projection (Ours) & \textbf{1.3134} & \textbf{1.1296} & \textbf{0.9468} & \textbf{1.0248} & \textbf{1.0370} & \textbf{1.4994} & \textbf{1.1790} & \textbf{1.0162} & \textbf{1.1470} & \textbf{1.0341} & \textbf{1.1344} & \textbf{1.0517} & \textbf{1.1261} \\
\midrule
\multirow{4}{*}{\textbf{MSE\textsuperscript{\(\downarrow\)}}}
&Tokenization        & 10.4342 & 18.5717 & 22.6030 & 19.5203 & 15.0847 & 19.5760 & 23.6075 & 16.1507 & 19.9191 & 17.5957 & 13.7098 & 22.7574 & 18.2942 \\
&Patch Reprogramming &  7.8320 & 21.9997 & 11.9019 &  9.1243 &  9.4781 & 17.4192 & 13.9960 & 10.2399 & 18.1153 & 10.3723 & 16.5644 &  6.3626 & 12.7838 \\
&Segment Embedding   & \underline{7.7900} & \underline{7.3849} & \underline{3.6623} & \underline{4.2979} & \underline{3.4580} & \underline{9.4851} & \underline{6.8501} & \underline{4.1459} & \underline{5.3810} & \underline{4.0613} & \underline{4.3267} & \underline{3.6439} & \underline{5.3739} \\
&Instance-level Projection (Ours) & \textbf{3.1676} & \textbf{3.0012} & \textbf{1.6113} & \textbf{2.3870} & \textbf{1.9398} & \textbf{6.5747} & \textbf{4.6274} & \textbf{2.2915} & \textbf{3.7346} & \textbf{2.6495} & \textbf{2.9348} & \textbf{2.5216} & \textbf{3.1201} \\
\midrule
\multirow{4}{*}{\textbf{R\(^2_{\Delta t}\)\textsuperscript{\(\uparrow\)}}}
&Tokenization        & -0.3638 & 0.2976 & 0.2377 & 0.2500 & 0.4291 & 0.3167 & 0.1206 & 0.4619 & 0.3507 & 0.4117 & 0.5114 & 0.1087 & 0.2610 \\
&Patch Reprogramming & -0.0236 & 0.1680 & 0.5986 & 0.6494 & 0.6413 & 0.3919 & 0.4786 & 0.6588 & 0.4095 & 0.6532 & 0.4097 & 0.7508 & 0.4822 \\
&Segment Embedding   & \underline{-0.0181} & \underline{0.7207} & \underline{0.8765} & \underline{0.8349} & \underline{0.8691} & \underline{0.6689} & \underline{0.7449} & \underline{0.8619} & \underline{0.8246} & \underline{0.8642} & \underline{0.8458} & \underline{0.8573} & \underline{0.7459} \\
&Instance-level Projection (Ours) & \textbf{0.5860}  & \textbf{0.8865} & \textbf{0.9457} & \textbf{0.9083} & \textbf{0.9266} & \textbf{0.7705} & \textbf{0.8276} & \textbf{0.9236} & \textbf{0.8783} & \textbf{0.9114} & \textbf{0.8954} & \textbf{0.9012} & \textbf{0.8634} \\
\bottomrule
\multicolumn{15}{l}{\footnotesize \textsuperscript{\(\downarrow\)} Lower is better, \textsuperscript{\(\uparrow\)} Higher is better; LLM4Delay is equipped with TS2Vec. All adaptation techniques here were implemented with Pythia-1B.}
\label{tab:fusioncompare}
\end{tabular}
}
\end{minipage}
}
\end{table*}
\clearpage

\begin{table}[h]
\centering
\caption{Trajectory Encoders Ablation Results}
{\setlength{\tabcolsep}{1pt}
\begin{tabular}{lcccccc}
\toprule
$f_{\text{enc}}(\cdot)$
& \textbf{MAE\textsuperscript{\(\downarrow\)}} &
& \textbf{MSE\textsuperscript{\(\downarrow\)}} &
& \textbf{R\(^2_{\Delta t}\)\textsuperscript{\(\uparrow\)}} & \\
\midrule
TS2Vec & 1.1261 & & 3.1201 & & 0.8634 & \\
\midrule
TCN-AE & 1.1232 & (-0.0029) & 3.2982 & (+0.1781) & 0.8501 & (-0.0133) \\
InfoTS & 1.2160 & (+0.0899) & 3.5050 & (+0.3849) & 0.8427 & (-0.0207) \\
ATSCC  & 1.4118 & (+0.2857) & 4.5745 & (+1.4544) & 0.7994 & (-0.0640) \\
\bottomrule
\multicolumn{7}{l}{\footnotesize \textsuperscript{\(\downarrow\)} Lower is better, \textsuperscript{\(\uparrow\)} Higher is better.} 
\end{tabular}
}
\label{tab:encablation}
\end{table}

According to Table~\ref{tab:encablation}, \(f_{\text{enc}}(\cdot)\) trained with TS2Vec achieves the best performance for our delay prediction formulation, while TCN-AE and InfoTS perform comparably with only slight degradation, and ATSCC, despite exhibiting strong class separability on the pretraining data, does not translate this into improved effectiveness. This experiment highlights the need for effective instance-level trajectory representations, especially when \(f_{\text{enc}}(\cdot)\) is frozen to reduce computational cost.

\subsection{Ablation Study of LLM Backbone}
To justify our choice of LLM backbone, this section evaluates our framework with varying $f_{\text{llm}}(\cdot)$, following the aforementioned evaluation protocol. Table~\ref{tab:llmablation} lists the evaluated LLM backbones and their average performance. Pythia-1B achieves the best overall performance, suggesting it is a promising backbone despite its small scale, potentially due to its pretraining corpus~\cite{pile} aiding generalization to aeronautical information. Under our setting and at this scale, English-centric models such as Pythia~\cite{pythia} and LLaMA~\cite{llama3} tend to outperform multilingual-focused models such as the Qwen3~\cite{qwen3.0} series, while instruction tuning has minimal impact on prediction accuracy.

\begin{table}[h]
\centering
\caption{LLM Backbones Ablation Results}
{\setlength{\tabcolsep}{1pt}
\begin{tabular}{lcccccc}
\toprule
$f_{\text{llm}}(\cdot)$
& \textbf{MAE\textsuperscript{\(\downarrow\)}}& 
& \textbf{MSE\textsuperscript{\(\downarrow\)}}&
& \textbf{R\(^2_{\Delta t}\)\textsuperscript{\(\uparrow\)}} &\\
\midrule
Pythia-1B & 1.1261 & & 3.1201 & & 0.8634 & \\
\midrule
LLaMA3.2-1B          & 1.4092 & (+0.2831) & 4.4818 & (+1.3617) & 0.7972 & (-0.0662) \\
LLaMA3.2-1B-Instruct & 1.3187 & (+0.1926) & 4.1222 & (+1.0021) & 0.8214 & (-0.0420) \\
Qwen3-0.6B-Base      & 1.4279 & (+0.3018) & 4.5710 & (+1.4509) & 0.8005 & (-0.0629) \\
Qwen3-0.6B           & 1.5252 & (+0.3991) & 5.0635 & (+1.9434) & 0.7835 & (-0.0799) \\               
\bottomrule
\multicolumn{7}{l}{\footnotesize \textsuperscript{\(\downarrow\)} Lower is better, \textsuperscript{\(\uparrow\)} Higher is better}
\end{tabular}
}
\label{tab:llmablation}
\end{table}

Although all models are provided with full context and trained to align trajectory embeddings with the LLM embedding space, different LLM backbones still exhibit varying performance. This highlights that accurate \(\Delta t_{\text{post},i}\) prediction depends not only on comprehensive inputs and effective cross-modality adaptation, but also on the model's linguistic capability to interpret them. LLM4Delay is compatible with arbitrary LLM backbones, demonstrating its scalability to larger, more capable models.

\subsection{Ablation Study on Context Inclusion}
A key advantage of LLM4Delay is its ability to flexibly integrate diverse operational information without fixed-dimensional feature preparation, providing the model with rich context. Since these contextual components vary in importance, we assessed their contributions by ablating $s_{i,t}$, then retraining and evaluating with identical hyperparameters, using the best-performing TS2Vec encoder and Pythia-1B backbone configuration. This quantifies the performance degradation from removing each context type, providing empirical justification for the design of $s_{i,t}$. Results are reported in Table~\ref{tab:promptablation}.

\begin{table}[h]
\centering
\caption{Context Removal Ablation Results}
{\setlength{\tabcolsep}{1pt}
\begin{tabular}{lcccccc}
\toprule
\textbf{Input Configs} 
& \textbf{MAE\textsuperscript{\(\downarrow\)}} &
& \textbf{MSE\textsuperscript{\(\downarrow\)}} &
& \textbf{R\(^2_{\Delta t}\)\textsuperscript{\(\uparrow\)}} & \\
\midrule
Full Context & 1.1261 & & 3.1201 & & 0.8634 & \\
\midrule
\textit{Textual Data} &&&&&&\\
w/o Textual Info.      & 1.1694 & (+0.0433) & 3.4121 & (+0.2920) & 0.8428 & (-0.0206) \\
\quad w/o Flight Info. & 1.1576 & (+0.0315) & 3.3221 & (+0.2020) & 0.8531 & (-0.0103) \\
\quad w/o METAR        & 1.1571 & (+0.0310) & 3.2380 & (+0.1179) & 0.8564 & (-0.0070) \\
\quad w/o TAF          & 1.1491 & (+0.0230) & 3.3021 & (+0.1820) & 0.8527 & (-0.0107) \\
\quad w/o NOTAMs       & 1.1645 & (+0.0384) & 3.3191 & (+0.1990) & 0.8519 & (-0.0115) \\
\midrule
\textit{Time-Series Data} &&&&&&\\
w/o Trajectories         & 3.3257 & (+2.1996) & 19.2201 & (+16.1000) & 0.2528 & (-0.6106) \\
\quad w/o Focusing Traj. & 2.1272 & (+1.0011) & 9.7092  & (+6.5891)  & 0.5907 & (-0.2727) \\
\quad w/o Active Traj.   & 1.2075 & (+0.0814) & 3.6068  & (+0.4867)  & 0.8433 & (-0.0201) \\
\quad w/o Prior Traj.    & 1.1980 & (+0.0719) & 3.4082  & (+0.2881)  & 0.8504 & (-0.0130) \\
\bottomrule
\multicolumn{7}{l}{\footnotesize \textsuperscript{\(\downarrow\)} Lower is better, \textsuperscript{\(\uparrow\)} Higher is better.}
\end{tabular}
}
\label{tab:promptablation}
\end{table}
First, removing textual inputs degrades performance, indicating that accurate prediction of \(\Delta t_{\text{post},i}\) depends on context beyond trajectories. Excluding \(P^{\text{F}}_{i,t}\), \(P^{\text{M}}_t\), \(P^{\text{T}}_t\), and \(P^{\text{N}}_t\) individually results in only mild degradation: \(P^{\text{F}}_{i,t}\) provides general operational context, consistent with most prior studies; \(P^{\text{M}}_t\) and \(P^{\text{T}}_t\) capture meteorological conditions and forecasts; and \(P^{\text{N}}_t\) describes special airspace events and operational constraints. These results align with intuition and prior findings, in which factors such as adverse weather, precipitation, runway closures, and equipment unavailability are known contributors to delays~\cite {Mueller2002}, captured here through contextual inputs encoded directly by the LLM.

Removing all trajectories causes the largest performance drop, indicating that trajectories are more influential than textual components. Trajectories capture aircraft motion and surrounding airspace conditions that dominantly affect \(\Delta t_{\text{post},i}\), and their removal also eliminates the learnable portion of the context. Among trajectory types, removing \(X^{f}_{i,t}\) causes the greatest degradation, as it directly captures aircraft motion relevant to \(\Delta t_{\text{post},i}\). Removing \(X^{a}_{i,t}\) also degrades performance due to lost airspace congestion information, whereas removing \(X^{p}_{i,t}\) causes only minor degradation, as it primarily captures historical rather than current conditions.

These results indicate that performance benefits from rich context relevant to \(\Delta t_{\text{post},i}\). Although aeronautical data often involves long context that can degrade performance, providing full context is preferable, as LLMs can selectively ignore irrelevant tokens while leveraging useful information when needed. The results also suggest the framework remains robust even when certain contexts are unavailable. We conjecture that different context combinations induce distinguishable distributions of \(\mathbf{h}_{i,t}\) in the LLM output space, easing the regression head's task.

\subsection{Visualization of Embedding Importance}
\begin{figure*}
    \centering
    \includegraphics[width=\textwidth]{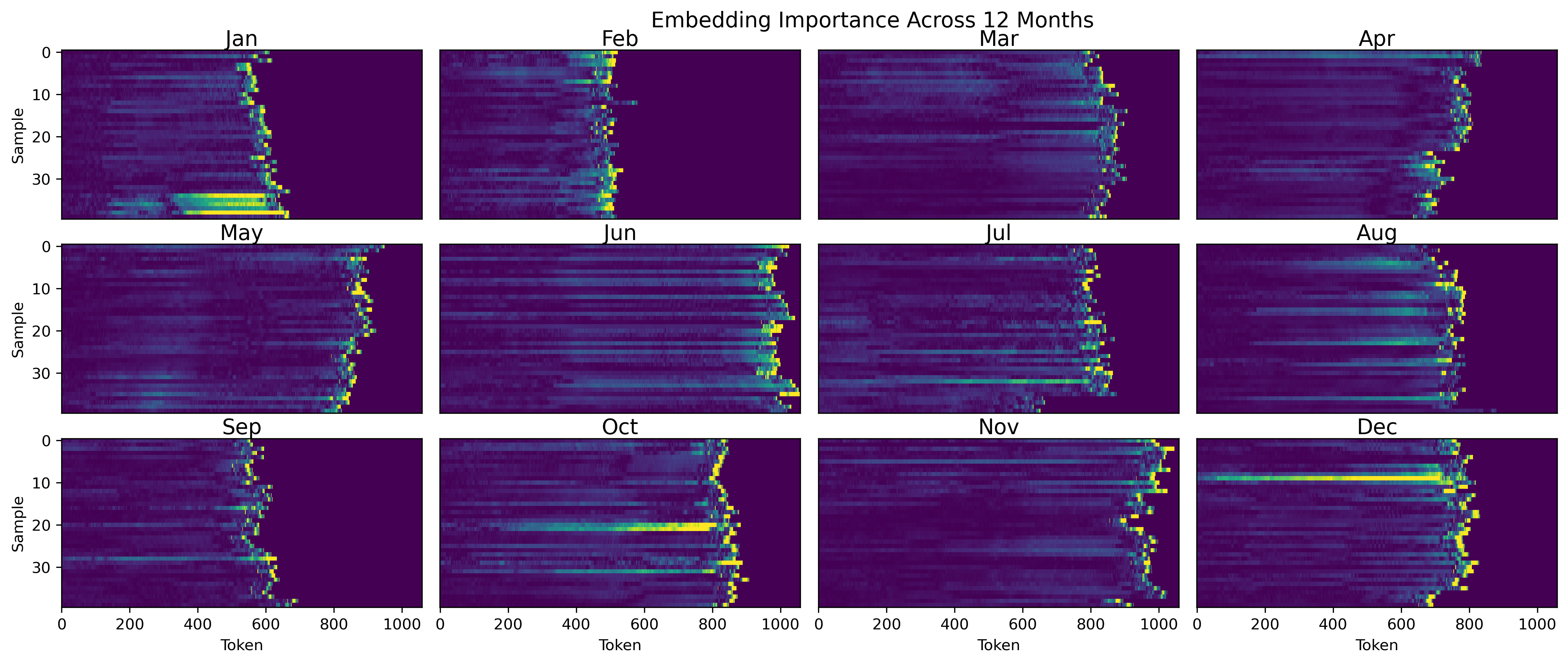}
    \vspace{-16pt}
    \caption{Token-Level Importance of Multimodal Prompt Components Across 12 Months}
    \label{fig:heatmap}
\end{figure*}

To further demonstrate model explainability, we visualize context importance using a perturbation-based sensitivity analysis, following prior LLM studies~\cite{importance_test}, in which each token in a sequence is perturbed, and the resulting change in output is measured. We extend this to our multimodal setting by perturbing the text-trajectory embedding sequence \(\mathbf{Z}_{i,t}\) index-by-index and observing the resulting shift in prediction, quantifying and visualizing the importance of each embedding. Since our task is formulated as regression, token importance is defined as the absolute change in predicted value induced by perturbation:
\begin{equation}
    I_{i,t/l} = \left| f_h(f_{\text{llm}}(\mathbf{Z}_{i,t/l}^{\text{pert}})[-1]) - f_h(f_{\text{llm}}(\mathbf{Z}_{i,t}^{\text{ori}})[-1]) \right|,
\end{equation}
for each index \(l\) in the input sequence \(\mathbf{Z}_{i,t}\). Each embedding is perturbed individually by replacing it with the LLM's pad token embedding, and the corresponding change in predicted \(\Delta t_{\text{post},i}\) is recorded. The magnitude of this change is visualized as a heat map, where lighter regions indicate greater importance. For each month, we sample 40 instances for visualization, with results shown in Fig.~\ref{fig:heatmap}.

In Fig.~\ref{fig:LLMxDelay}, the input sequence is constructed as \(\{ P^{\text{F}}_{i,t},P^{\text{M}}_{t},P^{\text{T}}_{t},P^{\text{N}}_{t},X^f_{i,t},X^a_{i,t},X^p_{i,t} \}\), with trajectory inputs occupying the latter portion. These positions show concentrated, high-importance regions, indicating that trajectory tokens contribute strongly to estimating \(\Delta t_{\text{post},i}\). Nevertheless, textual inputs remain relevant in certain samples, as indicated by localized regions of high sensitivity, suggesting that contextual text provides complementary information when needed.

\subsection{Model Demonstration}
\begin{figure*}
    \centering
    \subfloat[KAL672]{
        \includegraphics[width=0.5\textwidth]{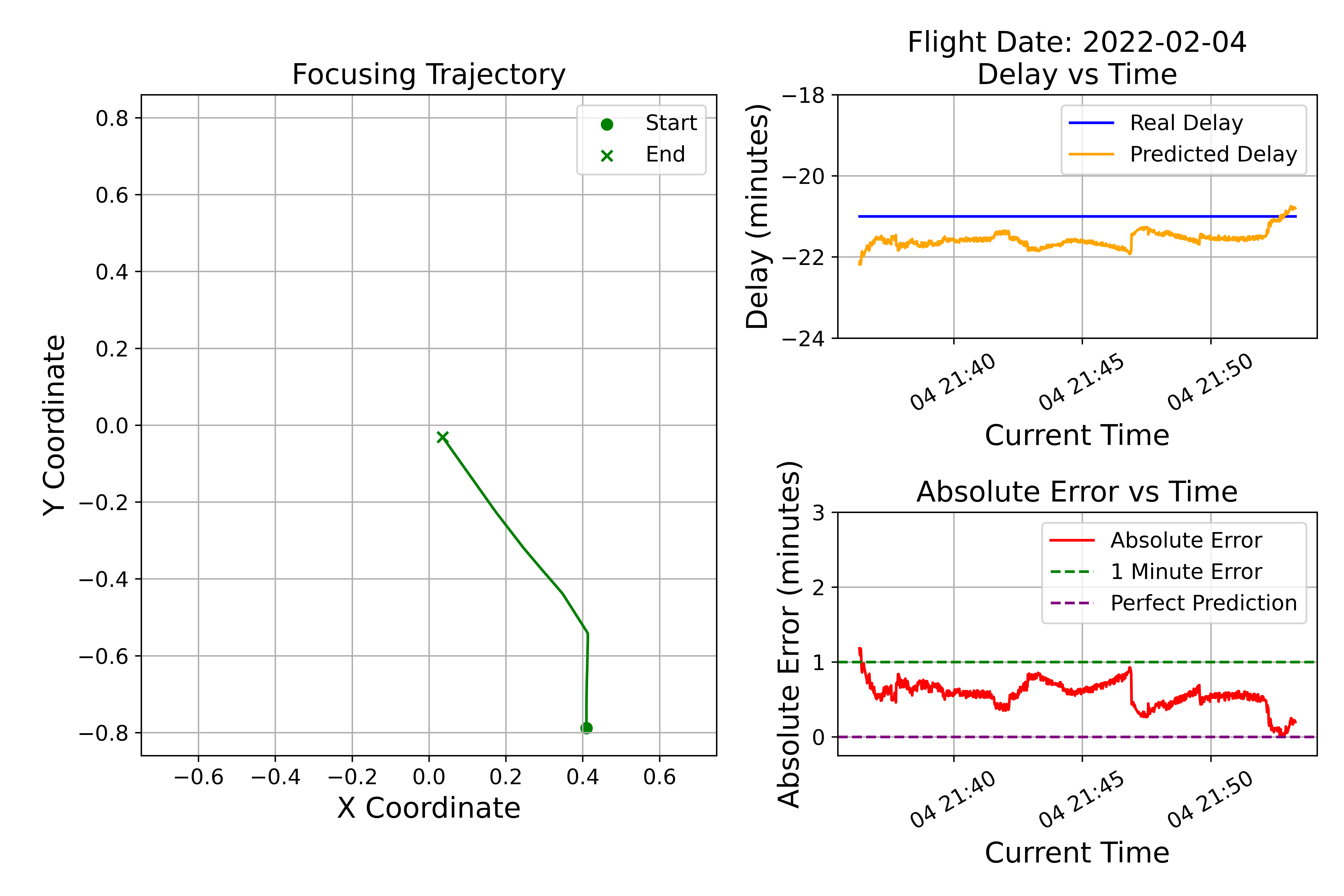}
        \label{fig:83408}
    }
    \subfloat[KAL742]{
        \includegraphics[width=0.5\textwidth]{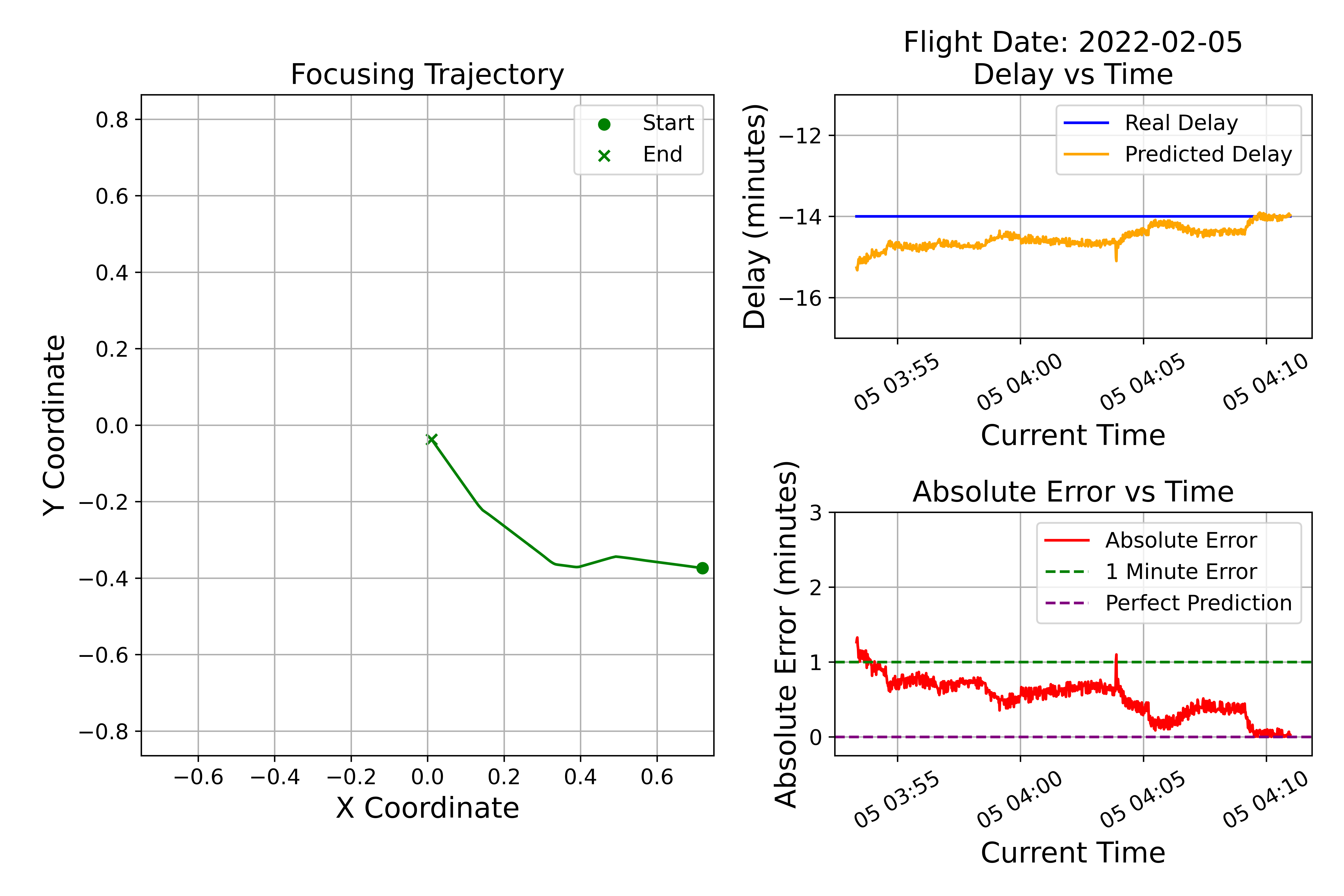}
        \label{fig:83451}
    }
    \caption{Demonstration of the LLM4Delay model. Each figure comprises three panels: (Left) full flight trajectory; (top right) predicted versus actual delay; and (bottom right) absolute prediction error over time.}
    \label{fig:realtimetraj}
\end{figure*}

We demonstrate our model's deployment to update delay predictions after aircraft enter the airspace, reflecting evolving context as new operational information becomes available to ATCs. This section adopts a monitoring-style setting where delay estimates are progressively updated as new information arrives. The model, trained on January data, was applied to sample flights from February, challenging it with entirely unseen data to demonstrate continuous prediction. For each sample flight $i$, full context $s_{i,t}$ was constructed at each time $t$, and the trained model updated $\hat{y}_{i,t} = f_{\theta}(s_{i,t})$ sequentially as new $s_{i,t}$ became available. Sample trajectories, predictions, and absolute errors over \([T_{entry}^{act},\, T_{\text{arrival}}^{act}]\) are visualized in Fig.~\ref{fig:realtimetraj}.

In these examples, the absolute error largely remains within 1 minute as predictions adapt over time, indicating progressive refinement. Since delays are recorded at the minute level in ATM~\cite{MLIT2024Airportal}, this error range indicates the framework's practical relevance for post-terminal monitoring. Furthermore, since arrival delays influence turnaround and departure operations~\cite{Wu01102003, Dnmez2024}, and ground-service scheduling depends on arrival-time information~\cite{Yan1998, Xu2024}, advance delay estimation can improve operational planning. Predicting \(\Delta t_{\text{post},i}\), the total time spent in terminal airspace and taxi-in, also supports monitoring of ICAO GANP KPIs such as KPI08 (Additional time in terminal airspace) and KPI13 (Taxi-in additional time)~\cite{ICAO9750}. Our framework thus enables advanced post-terminal delay estimation and proactive coordination of ground services, underscoring its operational and practical value.

\section{Conclusion}
This paper presents LLM4Delay, a framework that estimates delays by predicting post-terminal duration to support ATC operations at the destination airport. Unlike prior ATM approaches limited to fixed-size inputs or a single trajectory, LLM4Delay uses multimodal inputs, including aeronautical text and multiple trajectories, to better model complex delay-related factors. We propose instance-level projection, a cross-modality adaptation technique that projects instance-level trajectory embeddings into the LLM embedding space, outperforming prior techniques using fixed tokenization or feature-wise subseries encoding by enabling the LLM to directly capture instance-level trajectory semantics rather than inferring them from feature-wise segments. LLM4Delay leverages both a pretrained LLM and a trajectory encoder, whose complementary contributions are empirically validated, while updating only the adapter and regression head, enabling scalable replacement with higher-performing backbones. The framework demonstrates operational value for ATM by updating predictions as new information becomes available. Future work may extend the framework to other operational phases, explore broader ATM tasks beyond delay prediction, integrate additional modalities such as images or audio, and scale to higher-performing backbones.

\backmatter

\bmhead{Acknowledgements}
This work was supported by the Korea Agency for Infrastructure Technology Advancement (KAIA) Grant funded by the Ministry of Land, Infrastructure and Transport under Grant 22DATM-C163373-02. The work of Thaweerath Phisannupawong was supported by the Hyundai Motor Chung Mong-Koo Foundation Global Scholarship.

\bmhead{Conflict-of-Interest Statement}
On behalf of all authors, the corresponding author states that there is no conflict of interest.

\section*{Declarations}
\begin{itemize}
\item The code is publicly available at \href{https://github.com/petchthwr/LLM4Delay}{github.com/petchthwr/LLM4Delay}. 
\item The datasets are available at \href{https://huggingface.co/datasets/petchthwr/ICNDelay}{huggingface.co/datasets/petchthwr/ICNDelay}.
\end{itemize}









\bibliography{sn-bibliography}

\end{document}